\documentclass[accepted]{uai2022} % after acceptance, for a revised
\usepackage[american]{babel}
\usepackage{natbib} % has a nice set of citation styles and commands
\usepackage{mathtools} % amsmath with fixes and additions
\usepackage{booktabs} % commands to create good-looking tables
\usepackage{tikz} % nice language for creating drawings and diagrams

\usepackage[utf8]{inputenc} % allow utf-8 input
\usepackage[T1]{fontenc}    % use 8-bit T1 fonts
\usepackage{hyperref}       % hyperlinks
\usepackage{url}            % simple URL typesetting
\usepackage{booktabs}       % professional-quality tables
\usepackage{amsfonts}       % blackboard math symbols
\usepackage{nicefrac}       % compact symbols for 1/2, etc.
\usepackage{microtype}      % microtypography
\usepackage{xcolor}         % colors

\usepackage{amsmath}
\usepackage{amssymb}
\usepackage{graphicx}
\usepackage{caption}
\usepackage{subcaption}
\usepackage{dsfont}
\usepackage{wrapfig} % For inline figures
\usepackage{varwidth} % For placing figures next to tables
\usepackage{amsthm} % For theorems
\usepackage{thm-restate}
\usepackage{tcolorbox}

\newtheorem{definition}{Definition}

\definecolor{mydarkblue}{rgb}{0,0.08,0.45}
\hypersetup{ %
pdftitle={},
pdfauthor={},
pdfkeywords={},
pdfborder=0 0 0,
pdfpagemode=UseNone,
colorlinks=true,
linkcolor=red,
citecolor=blue,
filecolor=mydarkblue,
urlcolor=mydarkblue,
pdfview=FitH}

\usepackage[frozencache]{minted}
\newcommand{\ept}[1]{{\small \mintinline[breaklines,escapeinside=||,mathescape=true]{eptlexer.py:EPTLexer -x}{#1}}}

\definecolor{bluejl}{rgb}{0.0,0.6056031611752245,0.9786801175696073}
\definecolor{redjl}{rgb}{0.8888735002725198,0.43564919034818994,0.2781229361419438}
\definecolor{greenjl}{rgb}{0.2422242978521988,0.6432750931576305,0.3044486515341153}

\usepackage{multibib}
\newcites{supp}{References}

\title{Expectation Programming: Adapting Probabilistic Programming \\Systems to Estimate Expectations Efficiently}

\author[1]{Tim Reichelt}
\author[1]{Adam Goli{\'n}ski}
\author[1]{Luke Ong}
\author[1]{Tom Rainforth}
\affil[1]{%
    University of Oxford
}
  
\begin{document}
\maketitle

\begin{abstract}
We show that the standard computational pipeline of probabilistic programming systems (PPSs) can be inefficient for estimating expectations and introduce the concept of \emph{expectation programming} to address this.
In expectation programming, the aim of the backend inference engine is to directly estimate expected return values of programs, as opposed to approximating their conditional distributions.
This distinction, while subtle, allows us to achieve substantial performance improvements over the standard PPS computational pipeline by tailoring computation to the expectation we care about.
We realize a particular instance of our expectation programming concept, Expectation Programming in Turing (EPT), by extending the PPS \emph{Turing} to allow so-called \emph{target-aware} inference to be run automatically.
We then verify the statistical soundness of EPT theoretically, and show that it provides substantial empirical gains in practice.
\end{abstract}

\section{INTRODUCTION}

Estimating \emph{expectations} is at the center of many scientific workflows. 
For example, the decision theoretic foundations of most statistical paradigms,
e.g.~Bayesian decision theory, are rooted in calculating the expectation of a loss function~\citep{robert2004Monte}.

Carrying out this estimation often requires \emph{approximate inference} to be performed: we may not be able to directly draw samples of the random variable we wish to calculate the expectation of, or a simple Monte Carlo estimate might produce problematically high variance.

Probabilistic programming systems (PPSs) provide a powerful basis for encoding such inference problems and then assisting with, or even fully automating, the approximation of their solution~\citep{gordon2014Probabilistic,vandemeent2018Introduction}.
In a PPS, programs are typically specified (often indirectly) through an unnormalized density $\gamma(x)$.
Assuming analytic solutions are not available, the role of the system's inference engine is now to construct an approximation, $\hat{\pi}(x)$, for the distribution specified by the normalized density $\pi(x) = \gamma(x)/Z$, where $Z$ is an unknown normalizing constant and 
$\pi(x)$ typically represents a conditional distribution, such as the posterior in a Bayesian modeling setting.
This approximation can then be used in turn for downstream tasks,
such as approximating one or more expectations.

Though ostensibly very general, our key insight is that this standard PPS computational pipeline---which is implicitly followed by all contemporary PPSs that conduct inference approximately (e.g.~\citet{bingham2019Pyro,carpenter2017stan,cusumano-towner2019Gen,ge2018Turing,salvatier2016probabilistic,tran2016edward,wood2014New,mansinghka2014venture,goodman2014design,murray2018automated,minka2018infer})---can be highly suboptimal when our ultimate aim is to estimate a particular expectation, $\mathbb{E}_{\pi(x)}[f(x)]$.
This is because such a pipeline fails to perform estimation in a \emph{target-aware} fashion: it does not allow information about $f$ to be exploited by the inference engine, thereby forgoing the substantial empirical gains that using information about $f$ can yield~\citep{torrie1977nonphysical,hesterberg1988advances,wolpert1991monte,oh1992adaptive,evans1995methods,meng1996simulating,chen1997monte,gelman1998simulating,lacoste2011approximate,owen2013monte,golinski2019Amortizeda,rainforth2020Target}.
Note here that it is not generally possible to incorporate the required information about $f$ by adjusting the model definition; fundamental changes to the computational pipeline itself are required.

To address this, we introduce, and formalize, the concept of \emph{expectation programming}.
Here an expectation program is analogous to a probabilistic program, but its target quantity of interest is the expected value of the program's return values, rather than their conditional distribution.
This subtle distinction leads to changes in the requirements for the program to be valid, and, critically, the estimation that must be performed by the backend inference engine.
This, in turn, allows us to construct computational pipelines which are target-aware, utilizing information in the program itself to estimate expectations substantially more efficiently than can be achieved by existing PPSs.

We realize our expectation programming concept through a specific system we call 
\textbf{EPT} (Expectation Programming in Turing), built upon the Turing PPS \citep{ge2018Turing}. 
EPT takes as input a Turing-style program and uses a combination of program transformations and existing inference strategies to 
construct target-aware estimators via the TABI approach of~\citep{rainforth2020Target}.

We formally demonstrate the statistical soundness of EPT, proving that it produces consistent estimates under nominal assumptions.
We further show empirically that it can be used to express and run effective inference for a number of problems, finding that it produces  estimates that are significantly more accurate
than conventional usage of Turing.
As part of this, we also implement a new annealed importance sampling (AnIS) \citep{neal1998Annealed}
inference engine for Turing, finding that this allows for effective marginal likelihood estimation in a much wider
array of problems than Turing's previously supported inference strategies.

To summarize, our key contributions are: a) identifying the shortfall of existing PPSs when estimating expectations and introducing the concept of expectation programming to address this; b) developing EPT as a particular realization of the expectation programming concept; c) formalizing the notion of an expectation program and demonstrating the statistical correctness of EPT; d) introducing a new AnIS inference engine to Turing; and e) showing that EPT can provide substantial empirical benefits over conventional use of Turing on real problems.

\vspace{-3pt}
\section{BACKGROUND}
\vspace{-3pt}

\subsection{Turing Programs as Densities}
\label{sec:turing_details}

To provide a basis for introducing expectation programming, we consider the PPS Turing~(\citet{ge2018Turing},~{\small \url{https://turing.ml/dev/docs/using-turing/}}), but note that the concepts introduced apply to PPSs in general. 
We provide a brief introduction to Turing here,
along with our own new formalism for the densities Turing program define by extending the approach of~\citet[\S4.3]{rainforth2017Automating}. This is necessitated by some technical intricacies of the expectation programming approach.
To assist with this, we will use the following simple Turing program 
as a running example:
\vspace{-6pt}
\begin{minted}[breaklines,escapeinside=||,mathescape=true,numbersep=3pt,gobble=2,fontsize=\small]{eptlexer.py:EPTLexer -x} 
@model function model(y)
    x |$\sim$| Normal(0, 1)
    @addlogprob!(0.1)
    y |$\sim$| Normal(x, 1)
end
\end{minted}
\vspace{-6pt}
A Turing program is defined similarly to a normal Julia 
function~\citep{bezanson2017Julia}: the \ept{@model} macro indicates the 
definition of a Turing model, with tilde 
statements inside the body, e.g.
\ept{x |$\sim$| Normal(0, 1)}, to denote probabilistic model components. 
Observed data can be passed in as a formal argument to the function. 
If the variable name on the left-hand side of the tilde statement is not part 
of the arguments of the functions then it is interpreted as a random variable.

Let $x_{1:n}$ denote the set of direct outputs from sampling statements and
$y_{1:m}$ the observed data.
We can view 
Turing programs as defining an unnormalized density $\gamma(x_{1:n})$ (with an implicit appropriate
reference measure).
To compute the density for a given $x_{1:n}$ the program executes like a normal Julia program, while keeping track of the density of the current execution.
Specifically, when Turing reaches a tilde statement corresponding to a random variable, it 
samples a value for $x_i$, evaluates the density of this draw, and factors this
into the overall execution density.
We denote the density of the draw as $g_{i}(x_i|\eta_i)$, where $g_i$ denotes the form
of the sampling statement and $\eta_i$ its parameters.
For the tilde statements corresponding to the observed data,
it evaluates the density function $h_{j}(y_j|\phi_j)$---where 
$h_j$ and $\phi_j$ are analogous to $g_i$ and $\eta_i$ respectively---and factors the overall density accordingly.

Sometimes a user might want to add additional factors to the density without 
using a tilde statement. 
For this, Turing provides the \ept{@addlogprob!(log_p)} primitive which 
multiplies the density of the current execution by an arbitrary value \ept{exp(log_p)}.
We use $\psi_1, \dots, \psi_K$ to denote all the terms that are added to 
the density using \ept{@addlogprob!}.

Putting these together, the unnormalized density defined by any valid program trace can be written as 
\begin{equation}\label{eq:ppl_density}
  \gamma(x_{1:n}) = \prod_{i=1}^{n}g_{i}(x_i|\eta_i) \prod_{j=1}^{m}h_{j}(y_j|\phi_j) \prod_{k=1}^{K} \exp(\psi_k).
\end{equation}
Our example program thus defines the density 
$\gamma(x)\!=\!\exp(0.1) \mathcal{N}\!(x;0,1) \mathcal{N}\!(y;x,1),$ 
with a fixed input $y$.
Note here that everything (i.e. $n,x_{1:n},\eta_{1:n},g_{1:n},m,y_{1:m},\phi_{1:m},h_{1:m},K,\psi_{1:K}$) can be a random variable because of potential stochasticity in the program path.  
However, using the program itself, everything is deterministically calculable from $x_{1:n}$, which can thus be thought of as the `raw' random draws that dictate all the randomness of the program; everything else is a pushforward of these.

\vspace{-2pt}
\subsection{Target-Aware Inference}
\label{sec:tabi}
\vspace{-2pt}

Consider the problem of estimating an expectation of the form $\mathbb{E}_{\pi(x)}[f(x)]$ where $f(x)$ is known, but $\pi(x)$ cannot be directly evaluated or sampled from.
Namely, $\pi(x) = \gamma(x)/Z$ where $\gamma(x)$ is a known unnormalized density, but $Z$ is an unknown normalization constant (sometimes referred to as the marginal likelihood or partition function).

The inference engines in PPSs like Turing are setup to approximate $\pi(x)$ of this form.
As such, the standard pipeline to approximate an expectation using a PPS  is to first approximate $\pi(x)$ (e.g.~with samples)
and then use this to approximate the expectation in turn.

Unfortunately, this ignores information about $f$ and is therefore suboptimal if $f$ is known~\citep{golinski2019Amortizeda}.
While one might initially expect that information about $f$ can be easily incorporated through simple model adjustments, this is unfortunately not the case in practice: any adjustments we make will mean we need to estimate an additional corrective factor on top of performing inference for the new model.
Indeed, naive approaches to incorporating information about $f$, like adding $|f(x)|$ as a density factor to the model, have been found to typically worsen, rather than improve, the final estimates~\citep{rainforth2020Target}.

\citet{rainforth2020Target} recently showed that this issue stems from fundamental limitations of the efficacy of using a \emph{single} Monte Carlo estimator for such expectations.
Namely, through their Target-Aware Bayesian Inference (TABI) framework, they show 
that by breaking down the expectation into three parts:
\begin{equation}
  \label{eq:tabi}
  \mathbb{E}_{\pi(x)}[f(x)] = (Z^{+}_1 - Z^{-}_1)/Z_2,
\end{equation}
where 
$Z^{+}_1 = \int \gamma(x) f^{+}(x) dx$, $Z^{-}_1 = \int \gamma(x) f^{-}(x) dx$, $Z_2 =  \int \gamma(x) dx$, $f^{+}(x) = \max(f(x),0)$, and $f^{-}(x) = -\min(f(x),0)$,
and then estimating each term separately, one can often achieve a substantially improved overall estimator, 
$\mathbb{E}_{\pi(x)}[f(x)] \approx (\hat{Z}_1^+-\hat{Z}_1^-)/\hat{Z}_2.$

The intuition here is that each individual term can often be estimated more accurately in isolation than
the original expectation.
To see this, first note that the three subcomponents can be seen as the respective normalization constants 
of the three densities 
\begin{align}
\label{eq:gamma_def}
\begin{split}
\gamma_{1}^{+}(x) &\propto \gamma(x)f^{+}(x), \\
\gamma_{1}^{-}(x) &\propto \gamma(x)f^{-}(x), \\
\gamma_{2}(x) &= \gamma(x).
\end{split}
\end{align} 
The TABI framework now allows one to define a separate estimator \emph{tailored} to each of these problems.
In general, it allows one to repurpose any 
algorithm which provides estimates of the normalization constant into a 
target-aware inference algorithm by separately applying it to each of $\gamma_{1}^{+}(x)$, $\gamma_{1}^{-}(x)$,
and $\gamma_{2}(x)$.
TABI can theoretically achieve an arbitrarily low error 
for any fixed sample budget ($\ge3$),
unlike standard 
approaches such as self-normalized importance sampling or MCMC whose expected error is lower bounded, even when using an optimal proposal/sampler.
The achievable gains increase, both theoretically and empirically, with the degree of mismatch between $\pi(x)$ and $\pi(x)f(x)$.

\section{\hspace{-3pt}EXPECTATION PROGRAMMING}
\label{sec:methodology}

At a high level, \emph{expectation programming} adapts probabilistic programming systems to  
automate the \emph{estimation} of expectations in a \emph{target-aware} manner.
As we now explain, an \textit{expectation program} is analogous to a probabilistic program, but where the quantity
of interest is the expectation of its return values under the program's
conditional distribution, rather than the conditional distribution itself.

\subsection{Formalization}
To formalize the concept of an expectation program, we first statistically formalize probabilistic programs as follows.
\begin{definition}
\label{def:probabilistic_program}
A probabilistic program $\mathcal{P}$ in a probabilistic programming language defines an unnormalized density $\gamma(x_{1:n})$ over the raw random draws $x_{1:n} \in \mathcal{X}$ of the program, which collectively we refer to as the program trace,
along with an implicitly defined reference measure $\mu$. 
\vspace{-5pt}
\end{definition}

We let $\pi(x_{1:n}) = \gamma(x_{1:n}) / Z$ denote the normalized density with $Z = \int_\mathcal{X} \gamma(x_{1:n}) d\mu(x_{1:n})$. 
Here $\pi(x_{1:n})$ and $\mu$ combined implicitly define the conditional probability distribution specified by $\mathcal{P}$, which we denote $\mathbb{P}(A) = \int_A \pi(x_{1:n}) d\mu(x_{1:n})$.

To ensure that the induced probability measure of a program is well-defined, we require that $\gamma(x_{1:n})$ corresponds to a valid unnormalized density. 
This guarantees that there is a valid probability distribution the inference algorithm of the particular PPS can converge to.
We use this to formalize the concept of a \emph{valid} probabilistic program as follows.
\begin{definition}
\label{def:valid_probabilistic_program}
A probabilistic program, $\mathcal{P}$, is \emph{valid} (and defines a valid unnormalized probabilistic program density $\gamma(x_{1:n})$) if and only if both of the following hold: $\gamma(x_{1:n}) \geq 0, \forall x_{1:n} \in \mathcal{X}$; and $0 < \int_{\mathcal{X}}\gamma(x_{1:n}) d\mu(x_{1:n}) < \infty$.
\vspace{-5pt}
\end{definition}
For Turing we have described how programs specify $\gamma(x_{1:n})$ in Section~\ref{sec:turing_details}, but Definitions~\ref{def:probabilistic_program} and \ref{def:valid_probabilistic_program} apply more generally and only require that we can derive an unnormalized density function for a given program; a requirement that is satisfied by most existing popular PPSs.

We can now formalize the concept of an expectation program by associating return values to our program:
\begin{definition}
An expectation program, $\mathcal{E}$, is a probabilistic program (as per Definition~\ref{def:probabilistic_program}) with an associated set of return values $F\in \mathcal{F}\subseteq \mathbb{R}^d$ that are a deterministic mapping of the trace $x_{1:n}$.
\end{definition}
\vspace{-5pt}
From this definition we see that expectation programs are largely equivalent to probabilistic programs, indeed programs in any PPS that allows return values will also be expectation programs provided their outputs are numeric and fixed dimensional.
However, as their underlying quantity of interest is the expectation of their return values, $\mathbb{E}[F]$, they require a slightly different set of assumptions to ensure validity as follows.
\begin{definition}
\label{def:expectation_program}
An expectation program $\mathcal{E}$ is valid if and only if it is a valid probabilistic program and $F$ is integrable.
\end{definition}
\vspace{-5pt}
Here the additional requirement of the expectation program's outputs being integrable essentially equates to requiring that the expectation $\mathbb{E}[F]$ exists and $\mathbb{E}[|F_i|]<\infty$ for each dimension $F_i$ of $F$.
This is generally a very weak requirement, and strictly weaker than an assumption typically implicitly made by existing PPSs when confirming the validity of their inference engines as discussed in Appendix~\ref{apd:theory_details}.

To link expectation programs back into our early expectation notation, we now note that the requirement for the return values to be a deterministic mapping of the trace means that we can write $F=f(x_{1:n})$, such that $\mathbb{E}[F]=\mathbb{E}_{\pi(x_{1:n})}[f(x_{1:n})]$.
Thus the formal definition of the function we are taking the expectation 
of is that it is the full mapping from the raw random draws to the returned
values rather than what is lexically written 
in any \ept{return} statement(s). 
This is why, for instance, it is still valid to have multiple different \ept{return} 
statements in a program; provided each \ept{return} statement defines the same number of return 
values.  
In practice, this is not something we need to worry about when writing either models or inference engines as the law of the unconscious statistician relieves us from explicitly delineating the random variable defined by our function (the expectation of this random variable does not vary if we change the parameterization of our model).  
However, the distinction is important for ensuring validity and to identify the precise target function we wish to extract information about when making the inference target-aware.

\subsection{Target-Aware Inference Engines}

The key idea of our expectation programming paradigm is to use the formalisms from the previous section to set up inference engines that exploit information from $f$ to perform target-aware estimation.
As explained in Section~\ref{sec:tabi}, this can lead to estimators that provide substantial performance improvements over the standard PPS approach of simply approximating $\pi(x_{1:n})$, ignoring $f(x_{1:n})$ completely.

Note that the approximate computation we are performing here is fundamentally different to that of conventional inference engines: we are estimating an expectation, rather than approximating a conditional distribution.
This means the form of the outputs from our engine will change, while we will have to exploit additional information about the program.
As such, we will generally need to make changes to how the program itself is processed, rather than just implementing a new inference engine in the existing PPS structure.
Thankfully though, it will still usually be possible to repurpose existing inference engines as part of an overall target-aware estimation scheme, as we now show.

\subsection{Expectation Programming in Turing}

\begin{figure}[t]
\begin{minted}[breaklines,escapeinside=||,mathescape=true,numbersep=3pt,gobble=2,fontsize=\smaller]{eptlexer.py:EPTLexer -x} 
@expectation function expt_prog(y)
    x |$\sim$| Normal(0, 1)             # $x \sim \mathcal{N}(x; 0, 1)$
    y |$\sim$| Normal(x, 1)             # $y \sim \mathcal{N}(y; x, 1)$
    return x^3                  |$\phantom{\sim}$|# $f(x) = x^3$
end
expct_estimate, diagnostics = 
  estimate_expectation(expt_prog(2), 
    TABI(marginal_likelihood_estimator = 
      TuringAlgorithm(AnIS(),num_samples=100)))
\end{minted}
\vspace{-8pt}
\caption{
An example of estimating an expectation with EPT.
Here \ept{estimate_expectation} is our ``do estimation'' call which takes in expectation program 
\ept{expt_prog} (with input $y=2$) and an estimation method to apply (here a
TABI estimator using annealed importance sampling), and returns an estimate
for the expected return value of \ept{expt_prog}.
\vspace{-10pt}
}
\label{fig:ep_program}
\end{figure}

We now introduce a particular realization of the expectation programming concept which we call \emph{Expectation Programming in Turing} (EPT).
EPT builds on the PPS Turing to provide a highly effective, and surprisingly simple, mechanism to perform expectation programming.
It allows users to specify $\gamma(x)$ analogously to how they would using Turing's \ept{@model} macro, 
and uses Turing's \ept{return} semantics to define $F$ and thus $f(x)$.

The key component of the EPT is splitting up the estimation of the desired expectation
as per the TABI framework of Section~\ref{sec:tabi}.
To do so we use 
source-code transformations to generate three different Turing programs, one for 
each of the densities $\gamma_{1}^{+}(x)$, $\gamma_{1}^{-}(x)$, and 
$\gamma_{2}(x)$ (as per Equation~\eqref{eq:gamma_def}). We then estimate the expectation by individually estimating the 
normalization constant of each of these densities and then combining them as per 
Equation~\eqref{eq:tabi}.
Generating valid Turing programs 
allows us
to leverage any inference algorithm in Turing that 
provides marginal likelihood estimates to estimate the quantities $Z^{+}_1$,
$Z^{-}_1$, and $Z_2$. This modularity means that we do not have to implement 
custom inference algorithms that would only work with EPT.

Estimating expectations with EPT is done in two stages.
First, users define an expectation program
with the \ept{@expectation} macro, 
which is a  drop-in replacement for \ept{@model}, and an example for which is shown in Figure~\ref{fig:ep_program}. 
Using code transformations, \ept{@expectation} automatically generates the three Turing programs representing the 
densities $\gamma_{1}^{+}(x)$, $\gamma_{1}^{-}(x)$, and $\gamma_{2}(x)$.
This happens behind the scenes and the user does not need to deal with the transformed
programs directly.

\begin{figure*}[t]
\begin{minipage}[t]{0.5\textwidth}
\begin{minted}[breaklines,escapeinside=||,mathescape=true,numbersep=3pt,gobble=2,highlightlines={4},fontsize=\small]{eptlexer.py:EPTLexer -x} 
@expectation function expt_prog(y)
    x |$\sim$| Normal(0, 1) 
    y |$\sim$| Normal(x, 1)
    return x^3
end
\end{minted}
\end{minipage}
\begin{minipage}[t]{0.5\textwidth}
\begin{minted}[breaklines,escapeinside=||,mathescape=true,numbersep=3pt,gobble=2,highlightlines={4-6},fontsize=\small]{eptlexer.py:EPTLexer -x} 
@model function expt_prog(y)
    x |$\sim$| Normal(0, 1) 
    y |$\sim$| Normal(x, 1)
    tmp = x^3
    @addlogprob!(log(max(tmp, 0)))
    return tmp
end
\end{minted}
\end{minipage}
\vspace{-8pt}
\caption{
    The results of one of the three program transformations applied to the EPT \ept{@expectation} program from Figure~\ref{fig:ep_program} [left].
    Presented is the transformation into a valid Turing \ept{@model} program [right] 
    corresponding to the density $\gamma_{1}^{+}(x) \propto \gamma(x)f^{+}(x)$.  
    The transformed code fragment is highlighted. The full transformation is 
    slightly more complex due to Turing's internals. 
    Appendix~\ref{apd:macro_transformation} shows the full source code transformation for this model.
    \vspace{-12pt}
}
\label{lst:program_transformation}
\end{figure*}

To estimate the expectation,
the user then calls
\ept{estimate_expectation(expt_prog, method)}, where \ept{method} 
specifies the estimation approach to be used.
At present, the only supported class of methods is \ept{TABI}, which implements
the previously explained TABI estimators, but the syntax is designed to allow for easy 
addition of hypothetical alternative approaches. 

EPT then estimates
the normalization constants $Z^{+}_1$, $Z^{-}_1$, and $Z_2$ by running 
a Turing inference algorithm on each Turing program generated by \ept{@expectation} 
and combining the normalization constant estimates 
to form an estimate of the expectation. 
In the example in Figure~\ref{fig:ep_program}, we use \ept{TABI} with annealed importance sampling \ept{AnIS}, 
which is a new Turing inference algorithm that we have added to the system for the purposes of this paper.
\ept{TuringAlgorithm} is a thin-wrapper object
storing the necessary information that allows \ept{TABI} to use a Turing inference method.
\ept{AnIS} can be substituted with any other
Turing inference algorithm that returns a marginal likelihood estimate.
Here \ept{AnIS()} implies the use of some arbitrary default AnIS parameters
regarding the Markov chain transition kernel, and the number and
spacing of intermediate potentials used.

\vspace{-3pt}
\subsection{Program Transformations}
\label{sec:program_transformations}

We now consider how to generate the Turing programs corresponding to each of the TABI densities.
Note that expectation programs in EPT
are also valid Turing models, i.e., 
replacing \ept{@expectation} with \ept{@model} yields a valid Turing program.
Such a program corresponds to the unnormalized density $\gamma_{2}(x) \!=\! \gamma(x)$
without requiring any transformation of the source-code.

To create a Turing program corresponding to $\gamma_{1}^{+}(x)$, we need to 
multiply the unnormalized density of the unaltered Turing program $\gamma(x)$ by $\max(f(x),0)$. 
This is achieved using Turing's aforementioned \ept{@addlogprob!} primitive, such that we can think of it as adding a new  factor $\max(f(x_{1:n}),0)$ to the program density definition in~\eqref{eq:ppl_density}.
Our transformations are pattern matching procedures that find all the 
\ept{return expr} statements in the 
function body and then a) create a new local
variable \ept{tmp = expr} (where \ept{tmp} is a unique identifier generated using
\ept{gensym()}), b) insert a statement \ept{@addlogprob!(log(max(tmp, 0)))} 
before the \ept{return}, and c) change the return statement itself to
\ept{return tmp}. 
A concrete example of the transformation is presented in 
Figure~\ref{lst:program_transformation}.
The transformation for $\gamma_{1}^{-}(x)$ is analogous 
but inserts a statement \ept{@addlogprob!(log(-min(tmp, 0)))} instead.

Users can define multiple expectations by specifying multiple return values, while
each individual return value needs to almost surely be a numerical scalar.
This ensures that each target expectation is well defined and individually identified.
For each return expression, we apply our 
program transformation separately and derive a corresponding TABI estimator for each. For example, if we have 
\ept{return expr1, expr2, expr3}, the program transformation for 
$\{\gamma_1^+(x)\}_2$ would add the statement \ept{@addlogprob!(log(max(expr2, 0)))}.
Appendix~\ref{apd:multiple_expectations} shows a full example of this.

\vspace{-2pt}
\subsection{Validity of EPT}
\label{sec:validity}
\vspace{-2pt}

We now formalize and demonstrate the statistical correctness of the EPT approach.
For simplicity, we will assume throughout that programs almost surely return a single scalar value (i.e.~the probability that the return value fails to be a well-defined scalar is 0).
Generalization to programs with multiple return values is straightforward (provided the number of return values is fixed) by considering each return value separately in isolation (as EPT does itself).

\begin{restatable}{theorem}{tabi}
\label{thm:tabi_valid}
Let $\mathcal{E}$ be a valid expectation program in EPT with unnormalized density $\gamma(x_{1:n})$, defined on possible traces $x_{1:n}\in\mathcal{X}$, with return value $F=f(x_{1:n})$.
Then
$\gamma_1^+(x_{1:n}):=\gamma(x_{1:n})\max(0,f(x_{1:n}))$, $\gamma_1^-(x_{1:n}):=-\gamma(x_{1:n})\min(0,f(x_{1:n}))$, and $\gamma_2(x_{1:n}):=\gamma(x_{1:n})$ are all valid unnormalized probabilistic program densities.
Further, if $\{\hat{Z}_1^+\}_{m}$, $\{\hat{Z}_1^-\}_{m}$, $\{\hat{Z}_2\}_{m}$ are sequences of estimators for $m \in \mathbb{N}^+$ such that
\vspace{-5pt}
\begin{align*}
\{\hat{Z}_1^\pm\}_{m} &\overset{p}{\to}
\int_\mathcal{X} \gamma^\pm_1(x_{1:n}) d\mu(x_{1:n}),
\\ % \quad
\{\hat{Z}_2\}_{m} &\overset{p}{\to}
\int_\mathcal{X} \gamma_2(x_{1:n}) d\mu(x_{1:n})
\end{align*}

\vspace{-15pt}

where $\overset{p}{\to}$ means convergence in probability as $m\to \infty$, then
$(\{\hat{Z}_1^+\}_{m}-\{\hat{Z}_1^-\}_{m})/\{\hat{Z}_2\}_{m} \overset{p}{\to} \mathbb{E}[F].$
\end{restatable}

\vspace{-5pt}

Theorem~\ref{thm:tabi_valid}, which is proved in Appendix~\ref{apd:theory_details}, shows that if we have programs with the desired densities and we use consistent marginal likelihood estimators for each, then our resulting expectation estimates will themselves be consistent.
The latter is covered by the consistency of Turing's own inference engines.
The former requires that our transformed programs are valid Turing 
programs with the intended densities.  We now show that this is indeed the case.

Given an input EPT program $\mathcal{E}$, EPT applies transformations 
to get the three Turing programs $\mathcal{P}_1^+$, $\mathcal{P}_1^-$, and $\mathcal{P}_2$ with  $\gamma_1^+(x_{1:n})$, $\gamma_1^-(x_{1:n})$, and 
$\gamma_2(x_{1:n})$ as their respective densities.
To ensure that the transformations for $\gamma_1^+(x_{1:n})$ and $\gamma_1^-(x_{1:n})$
are correct, we need to ensure that a) the inserted code
in our transformations is itself valid, 
b) the transformation does not have any
unintended side effects, and 
c) the new density terms add valid factors to the program
density.  
The first is true as the operation of the
transformed sections of code are identical to the originals except for
the new \ept{@addlogprob!} terms, which themselves produce no outputs and, by 
construction, use only the variables that are in scope.
The second is guaranteed by ensuring that the \ept{tmp} variables are given unique
identifiers that cannot clash with each other or any other variables in the program.
The third follows from the restriction that each return value must almost surely
be a numerical
scalar, coupled with the fact that the added density factors (namely \ept{max(tmp, 0)} and \ept{-min(tmp, 0)}) are non-negative by construction.

Thus, we have shown that EPT will produce a consistent estimation of program expectations, under the assumptions of Definition~\ref{def:expectation_program} and the consistency of the base inference algorithms implemented in Turing.

\begin{figure*}[h!]
    \centering
    \begin{subfigure}{0.4\textwidth}
        \centering
        \includegraphics[clip,trim=1cm 0cm 0cm 0cm,width=\columnwidth]{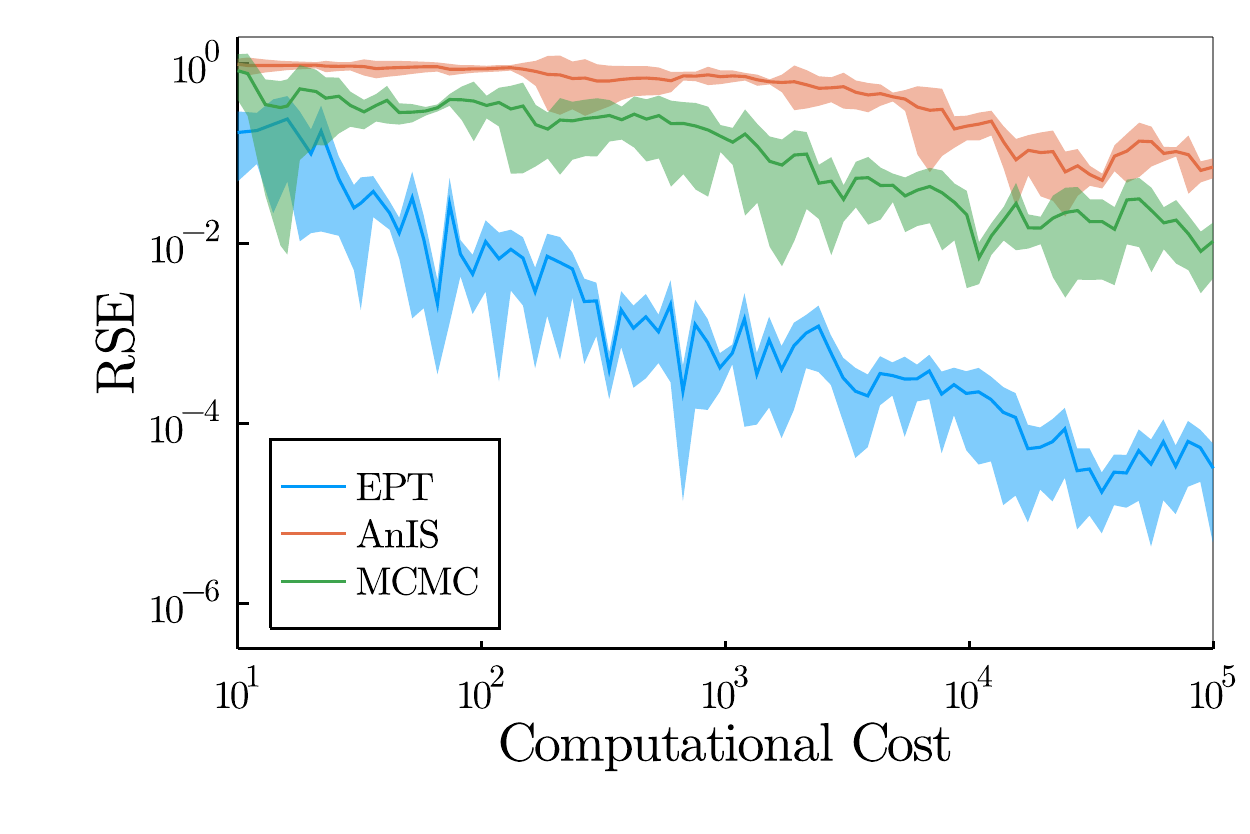}
    \end{subfigure}
    \qquad
    \begin{subfigure}{0.4\textwidth}
        \centering
        \includegraphics[clip,trim=1cm 0cm 0cm 0cm,width=\columnwidth]{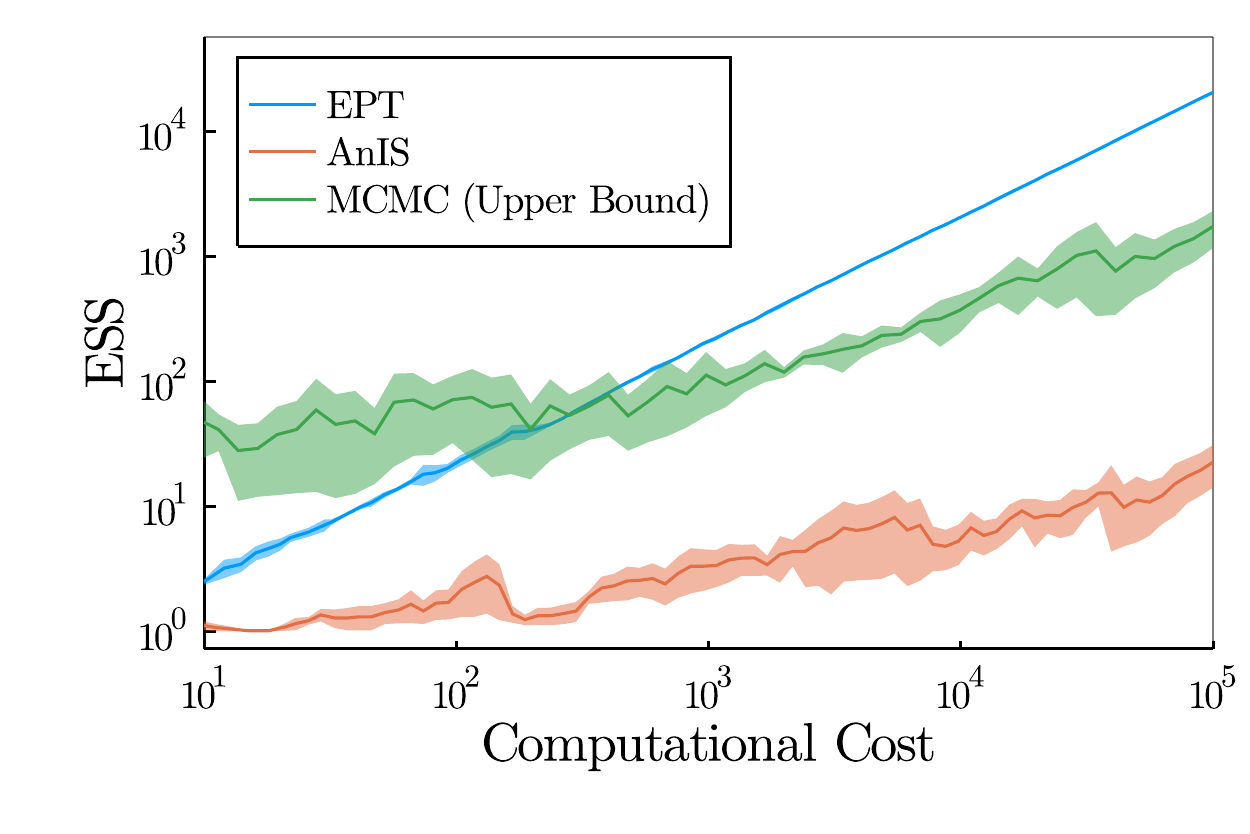}
    \end{subfigure}
    \vspace{-12pt}
    \caption{Relative squared error (RSE) and effective sample size (ESS) for the Gaussian posterior 
    predictive experiment for a given computational cost. 
    This cost is normalized across approaches by using the same number of likelihood evaluations and it has units of the combined number of samples used by EPT, which is equivalent to half the AnIS samples produced or $1/(2n)$ of the number of MCMC samples produced, where $n$ is the number of intermediary distributions used by AnIS.
    The solid lines show the 
    median of the estimator while the shaded region show the 25 \% and 75 \% quantiles.
    Medians and quantiles are computed over 10 separate runs with different random 
    seed for the posterior predictive problem. For the ESS plot we are plotting 
    $\min(\text{ESS}_{Z_1}, \text{ESS}_{Z_2})$; note that our estimates are (quite loose) upper bounds for MCMC (see text).
    }
	\vspace{-4pt}
    \label{fig:post_pred_experiment}
\end{figure*}

\begin{figure*}[h!]
    \centering
    \begin{subfigure}{0.4\textwidth}
        \centering
        \includegraphics[clip,trim=1.0cm 0cm 0cm 0cm,width=\columnwidth]{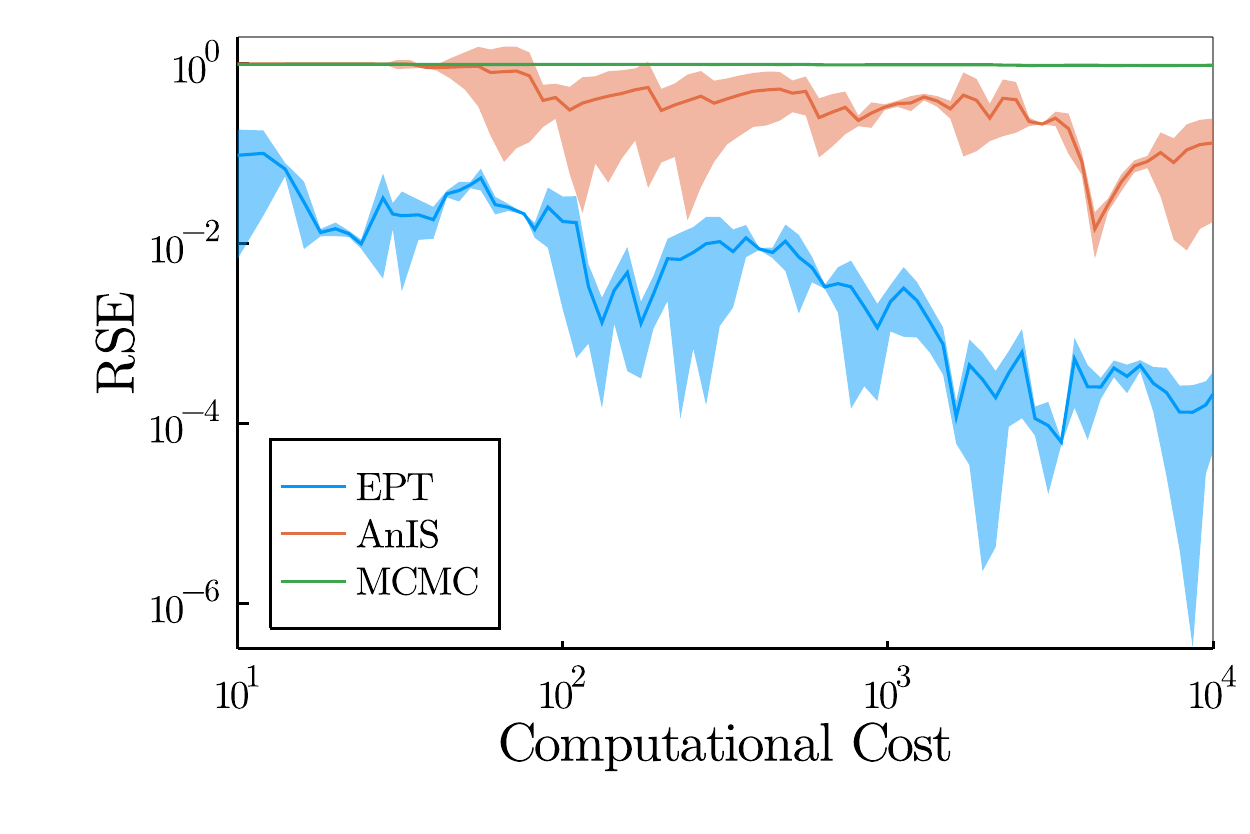}
    \end{subfigure}
    \qquad
    \begin{subfigure}{0.4\textwidth}
        \centering
        \includegraphics[clip,trim=1.0cm 0cm 0cm 0cm,width=\columnwidth]{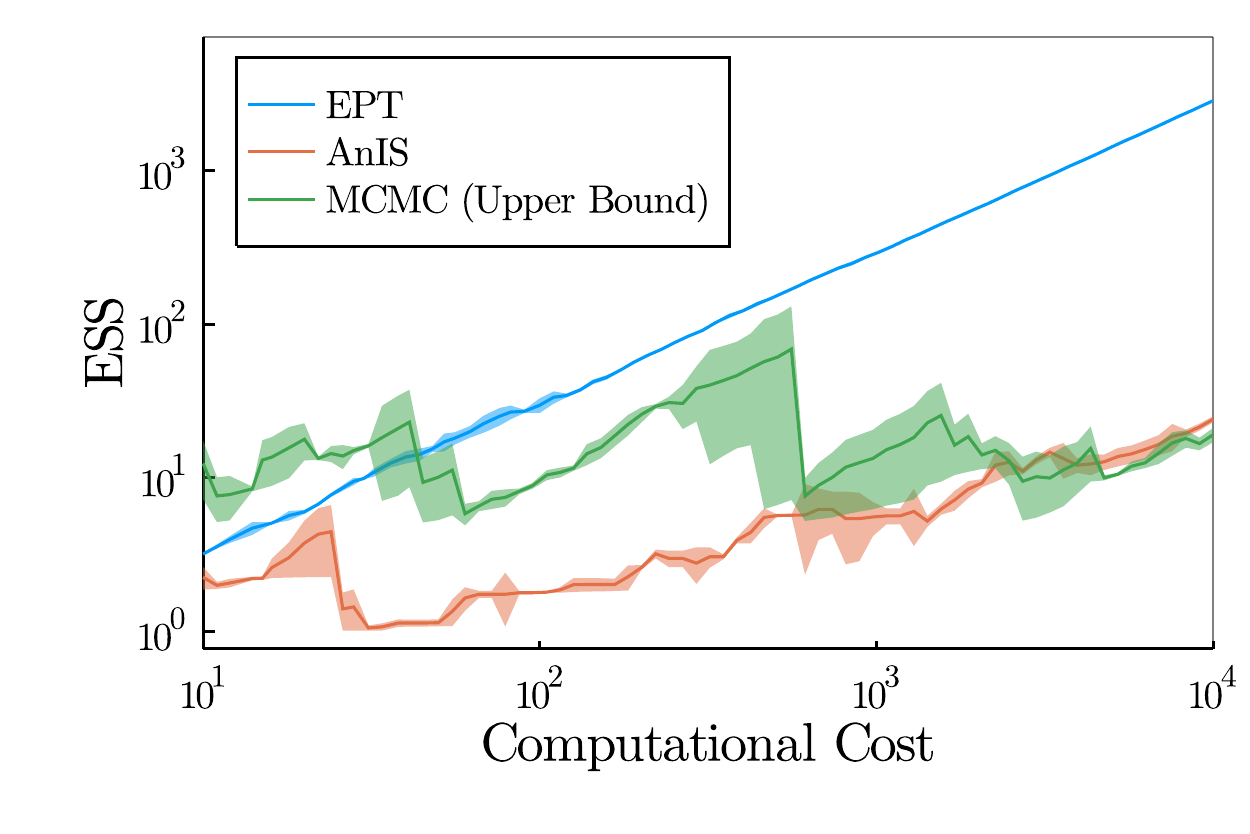}
    \end{subfigure}
    \vspace{-12pt}
    \caption{RSE and ESS for the SIR experiment. Conventions as in Figure~\ref{fig:post_pred_experiment}; results computed over 5 runs.}
	\vspace{-8pt}
    \label{fig:sir_experiment}
\end{figure*}

\vspace{-7pt}
\section{RELATED WORK}
\label{sec:related_work}
\vspace{-4pt}

Our focus is explicitly on the case of \emph{estimating} expectations.
Though a few papers \citep{gordon2014Probabilistic,zinkov2017Composinga} 
have provided alternative formalizations for the expectation defined by a 
probabilistic program, none do this from the perspective of directly targeting this expectation as the quantity to estimate.
Relatedly, a few languages provide primitives to compute expectations \emph{analytically} in the rare situation where this is possible, such as Hakaru \citep{zinkov2017Composinga} or $\lambda$PSI \citep{gehr2020lambdapsi}.
Unlike in our setting, these do not require notable changes to the backend computation from the standard inference setting because the underlying problem remains the same: calculate an integral analytically.
The contributions of these works are thus somewhat tangential to our own, with our key message being that \emph{estimating} expectations \emph{efficiently} requires a distinct computational pipeline to that of modern PPSs.

Some PPSs also provide syntactic sugars for forming expectation estimates from the samples
produced by inference, but these do not adjust the inference itself to exploit target function information.
For example,
in Stan \citep{carpenter2017stan} users can apply target 
functions to posterior samples using the \texttt{generated\_quantities} block. 
Similarly, in Pyro \citep{bingham2019Pyro} the return values are stored along with MCMC 
posterior samples, thus allowing expectations to be estimated by taking empirical averages. 
PyMC3 \citep{salvatier2016probabilistic}
allows users to track deterministic transformations of the latent variables. 
Turing itself also provides a \texttt{generated\_quantities} function, similar to Stan (see Appendix~\ref{apd:turing_expectation} for an example).

\section{EXPERIMENTS}
\label{sec:experiments}
\vspace{-6pt}

We demonstrate the effectiveness of the EPT target-aware inference methods on three problems: a synthetic 
numerical example, an SIR epidemiology model, and a Bayesian hierarchical model.
Our EPT implementation and the code for all experiments can be found at \url{git.io/JZOqN}.

The performance of EPT depends on the performance of the chosen marginal likelihood estimator.
At the time of writing, Turing provides implementations of Sequential Monte Carlo 
\citep{del2006sequential} and Importance Sampling (IS) as 
inference algorithms that provide marginal likelihood estimates, but only allows using the prior as the proposal which can never be target-aware. 
To address this issue, we implemented a new Turing inference engine that uses Annealed Importance Sampling (AnIS) \citep{neal1998Annealed} (see Appendix~\ref{apd:anis}), chosen because of its ability to estimate normalization constants in high dimensions \citep{wallach2009evaluation,salakhutdinov2010efficient,Wu2017on}.

AnIS requires setting two hyperparameters: an annealing schedule and a transition kernel. Currently,
users can choose between two transition kernels: 
Metropolis-Hastings (MH) 
implemented in \texttt{AdvancedMH.jl} \citep{turingdevelopmentteam2020TuringLang}
and Hamiltonian Monte Carlo (HMC) \citep{neal2011MCMC,hoffman2014TheNS,betancourt2018conceptual} 
in \texttt{AdvancedHMC.jl} \citep{xu2020AdvancedHMC}.
To ensure a fair 
comparison we use the same setup and hyperparameters for
both \textcolor{bluejl}{EPT}'s backend and standard, non-target-aware \textcolor{redjl}{AnIS}.
We also compare directly to \textcolor{greenjl}{MCMC} targeting the posterior and
using the same type of transition kernel as \textcolor{redjl}{AnIS} and \textcolor{bluejl}{EPT}.
This transition kernel is MH in Section~\ref{sec:exp_post_pre} and HMC elsewhere.
Detailed configurations are given in
Appendix~\ref{apd:exp_hyperparams}.

To compare the performance of the estimators
we look at the effective sample size (ESS, see below) and the relative squared error (RSE) $\hat{\delta} := (\hat{\mu} - \mu)^2/\mu^2$, where
$\mu$ denotes the ground-truth value and $\hat{\mu}$ is the estimate. 
All our experiments correspond to target functions which are always positive, so we use $Z_1$ to refer to $Z^{+}_1$ as $Z^{-}_1 = 0$.
Appendix~\ref{apd:positive_target_function} shows how EPT can avoid computation for $Z^{-}_1$ when possible. 

Both EPT and AnIS produce 
weighted samples $\{w_{\ell},\hat{x}_{1:n}^{\ell}\}_{\ell}$, so we use
$\text{ESS}(\{w_{\ell},\hat{x}_{1:n}^{\ell}\}_{\ell})\!=\!(\sum_\ell w_\ell)^2 / \sum_\ell w_\ell^2$.
EPT produces two sets of samples (for $Z_1$ and $Z_2$ respectively), so we take our overall ESS as $\min(\text{ESS}_{Z_1},\text{ESS}_{Z_2})$.
For AnIS, we only produce one set of samples (targeting $Z_2$) but use them to estimate both $Z_1$ and $Z_2$.
Here $\text{ESS}_{Z_2}^{\text{AnIS}}$ can be calculated in the normal way, but we have
$\text{ESS}_{Z_1}^{\text{AnIS}}(\{w_{\ell},\hat{x}_{1:n}^{\ell}\}_{\ell})= (\sum_\ell w_\ell f(\hat{x}_{1:n}^{\ell}))^2 / \sum_\ell (w_\ell f(\hat{x}_{1:n}^{\ell}))^2$.
As MCMC produces unweighted samples, we cannot directly calculate analogous ESSs.
Instead, we calculate an upper bound on the true ESS by assuming that the autocorrelation between samples is zero, i.e. that samples are independent.
$\text{ESS}_{Z_2}^{\text{MCMC}}$ is then just equal to the number of samples produced, while $\text{ESS}_{Z_1}^{\text{MCMC}}(\{\hat{x}_{1:n}^{\ell}\}_{\ell})= (\sum_\ell f(\hat{x}_{1:n}^{\ell}))^2 / \sum_\ell (f(\hat{x}_{1:n}^{\ell}))^2$.

\vspace{-3pt}
\subsection{Gaussian Posterior Predictive}
\label{sec:exp_post_pre}
\vspace{-3pt}

The first problem considered is calculating the posterior predictive distribution of a Gaussian model with an unknown mean, where
$\gamma(\mathbf{x})=\mathcal{N}(\mathbf{x}; 0, I) \,\mathcal{N}(\mathbf{y}; \mathbf{x}, I)$ and $f(\mathbf{x})=\mathcal{N}(-\mathbf{y}; \mathbf{x}, \frac{1}{2}I)$
are the unnormalised density and target function, respectively.
We assume our observed data is $\mathbf{y} =  (3.5/\sqrt{10}) \mathbf{1}$ where 
$\mathbf{1}$ is a 10-dimensional vector of ones.
Using EPT we can express this expectation in just 5 lines of code---the full model 
is given in Appendix~\ref{apd:post_pred_ept}.
This problem is amenable to an analytic solution so allows us to compute the error of the estimates.
Figure~\ref{fig:post_pred_experiment} compares the performance of \textcolor{bluejl}{EPT}, \textcolor{redjl}{AnIS}, 
and \textcolor{greenjl}{MCMC} (here MH).
We see a clear benefit 
to using the target-aware inference algorithm to estimate the expectation.
EPT achieves a lower RSE, and the ESS highlights the advantage
of using separate estimators for $Z_1$ and $Z_2$.
Note that the high apparent ESS of MCMC for small sample sizes is likely due to the looseness of the bound, rather than the true actual ESS being large.  

\vspace{-3pt}
\subsection{SIR Epidemiological Model}
\vspace{-3pt}

Our second problem setting is 
a more applied example based on the Susceptible-Infected-Recovered (SIR)
model of \citet{kermack1927contribution} from the field of epidemiology. Assume we face
a disease outbreak.
The government has provided us with a function yielding the expected cost of the disease 
which depends on the basic reproduction rate
$R_0$, which indicates the expected number of people one infected person will infect 
in a population where everyone is susceptible.
We seek to infer $R_0$ and the expected cost of the outbreak. 

The SIR model divides the population into three compartments: 
people who are susceptible to the disease,
those who are currently infected, 
and those who have already recovered. 
The dynamics of the outbreak are modelled by a set of differential equations 
\begin{align}
    \label{eq:sir_diffeq}
    \frac{dS}{dt} = - \beta S \frac{I}{N}, \quad 
    \frac{dI}{dt} = \beta S \frac{I}{N} - \gamma I, \quad
    \frac{dR}{dt} = \gamma I,
\end{align}
with parameters $\beta$ and $\gamma$. $S$, $I$ and $R$ correspond to the number of people susceptible, infected and 
recovered, respectively. The size of the total population is $N = S+I+R$.
Roughly, $\beta$ models the constant rate of infectious contact between people, 
while $\gamma$ is the constant recovery rate of infected individuals. From these 
parameters we can calculate the basic reproduction rate
$R_0 = \beta/\gamma$. 
We assume $\gamma$
to be known, and we want to infer $\beta$ and the initial number of infected people 
$I_0$.
The full statistical model and the cost function (which is based on $R_0$) 
is given in Appendix~\ref{apd:sir_exp}.

This scenario is a good use case for EPT because we are interested in estimating 
a specific expectation with high accuracy. 
Our cost function has some outcomes which might have low probability under the posterior but which incur a very high cost. 
These outcomes are liable to be missed by non-target-aware schemes, leading to extremely
skew estimators that almost always underestimate the expectation.

Figure~\ref{fig:sir_experiment} compares the performance of the 
estimators. 
Since this problem is not amenable to an analytic solution, we estimate the ground-truth using
a customized IS estimator with orders of magnitude more samples than estimates presented in the plot (see
Appendix~\ref{apd:exp_hyperparams}).
EPT substantially improves on the baselines, with \textcolor{greenjl}{MCMC} (here HMC) failing to provide
any meaningful estimate; it produces no samples where $f(x)$ is significant.
\textcolor{bluejl}{EPT} is able to overcome this through its use of a separate estimator for $\gamma(x)f(x)$.
The fact that \textcolor{greenjl}{MCMC} does far worse than \textcolor{redjl}{AnIS}, despite neither being target-aware, stems from the latter producing
a greater diversity of (weighted) samples, a small number of which land in regions of high $f(x)$ by chance.
To confirm that the failure of MCMC is not due to the specific implementation used we also computed results for this model in Stan, which produced similar results, see Appendix~\ref{apd:sir_discussion}.

 \subsection{Hierarchical Concentration Model}
\label{sec:radon_experiments}

\begin{figure}[t]
\begin{center}
\vspace{-10pt}
\centerline{\includegraphics[clip,trim=1.0cm 0cm 0cm 0cm,width=0.4\textwidth]{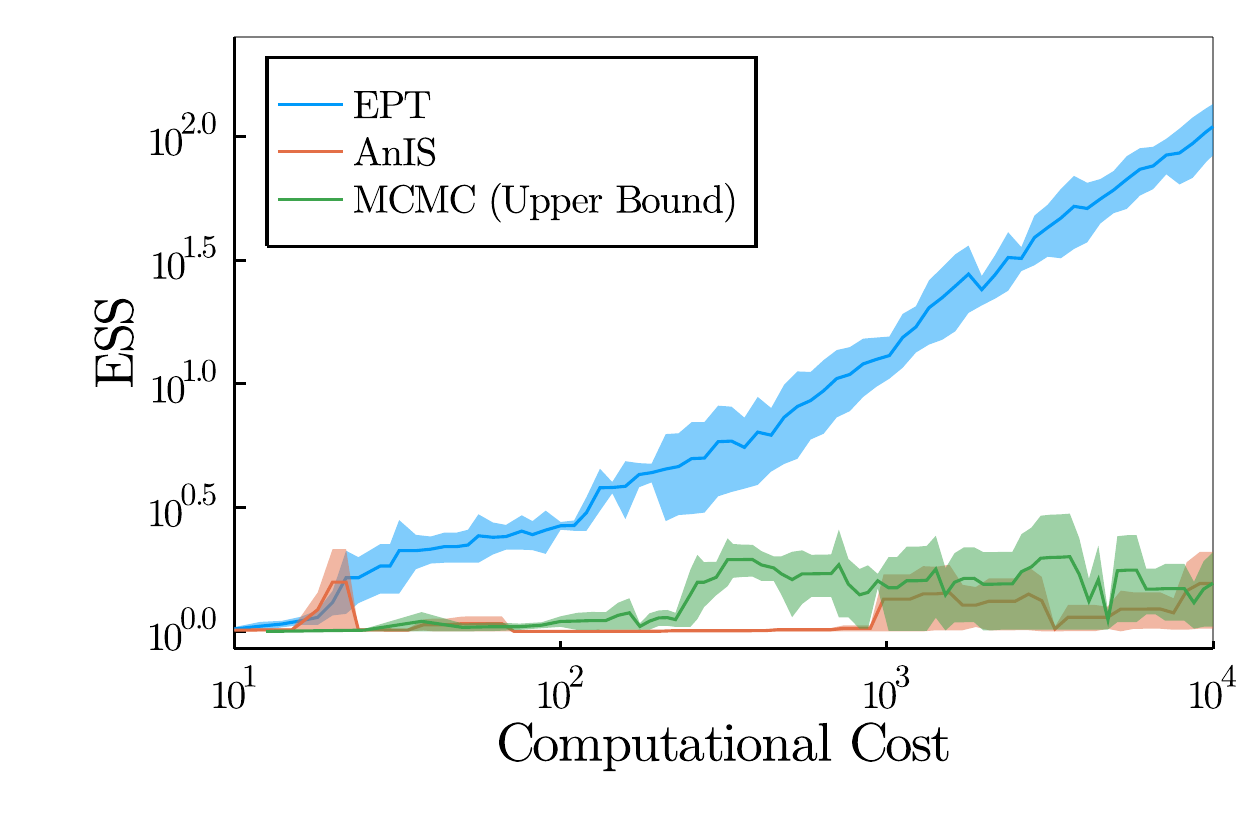}}
\vspace{-10pt}
\caption{ESS plots for the Radon experiment. Conventions as in 
Figure~\ref{fig:post_pred_experiment}; estimates based on 10 runs/seeds.
}
\vspace{-10pt}
\label{fig:radon-ess}
\end{center}
\end{figure}

Our third problem setting is a Bayesian hierarchical model for the radon concentration 
in households in different counties, adapted from \citet{gelman2006hill}.
For the $j^{\text{th}}$ house in county $i$, we would like to predict the log radon 
concentration $y_{ij}$ inside the house. For each house we have a covariate $x_{ij}$
which is $0$ if the house has a basement, and $1$ if it does not. With this setup, the model is defined as
\vspace{-2pt}
\label{eq:radon_equations}
\begin{align}
    \mu_{\alpha} &\sim \mathcal{N}(0, 10),
    &\alpha_i &\sim \mathcal{N}(\mu_{\alpha}, 0.12), \displaybreak[0] \\
    \mu_{\beta} &\sim \mathcal{N}(0, 10), 
    &\beta_i &\sim \mathcal{N}(\mu_{\beta}, 0.22), \displaybreak[0] \\
    \epsilon &\sim \text{HalfCauchy}(0, 5), %\nonumber\\
    &y_{ij} &\sim \mathcal{N}(\alpha_i + \beta_i x_{ij}, \epsilon). \label{eq:radon_pred}
\end{align}
We now want to find out whether the radon level in \emph{all} households
is below an acceptable level, taking this threshold to be 4pCi/L.
The probability of 
this event is equal to the expectation under the posterior of a step function $f(x)$.
However, to allow the use of HMC transition kernels we use a logistic function as a continuous relaxation of this step function. 
See Appendix~\ref{apd:radon_exp} for more details.

\begin{table}[t]
    \vspace{-2pt}
	\caption{Final estimates for the Radon experiments. Mean and 
		standard deviation estimated over 10 runs.}
    \vspace{-10pt}
	\label{tab:radon-estimates}
	\begin{center}
		\begin{small}
			\begin{sc}
				\begin{tabular}{lcccr}
					\toprule
					Method & Final Estimate  \\
					\midrule
					\textcolor{bluejl}{EPT}    & $3.74\mathrm{e}{-8} \, \pm \, 2.39\mathrm{e}{-9}$ \\
					\textcolor{redjl}{AnIS}      & $1.15\mathrm{e}{-9} \, \pm \, 3.02\mathrm{e}{-9}$ \\
					\textcolor{greenjl}{MCMC}      & $7.79\mathrm{e}{-18} \, \pm \, 2.46\mathrm{e}{-17}$ \\
					\bottomrule
				\end{tabular}
			\end{sc}
		\end{small}
	\end{center}
\end{table}

This problem cannot be solved analytically and estimating the ground-truth with sufficient accuracy is computationally infeasible. 
We, therefore, resort to comparing \textcolor{bluejl}{EPT} and \textcolor{redjl}{AnIS} based on their ESSs, noting that a low ESS almost
exclusively means a poor inference estimate, while a high ESS is a strong (but not absolute) indicator of good
performance.
As we can see in Figure~\ref{fig:radon-ess}, \textcolor{bluejl}{EPT} outperforms 
standard \textcolor{redjl}{AnIS} by several orders of magnitude. 
Additionally, Table~\ref{tab:radon-estimates}
presents the final expectation estimates for each method. All methods differ in 
their estimates.
However, \textcolor{bluejl}{EPT} is the only one where the standard deviation of the estimate is small relative to
its mean estimate, which, coupled with our ESS results, provides strong evidence that it is significantly
outperforming the baselines.
In particular, it seems clear that the \textcolor{greenjl}{MCMC} (here HMC) estimate is very poor: the fact that its estimate is many orders
of magnitude smaller than the others, coupled with its extremely low ESS (despite ignoring sample correlations), shows that it is failing to
produce any samples in regions where $f(x)$ is non-negligible.

\section{CONCLUSION}

We have introduced the concept of expectation programming which describes the process of encoding expectations programmatically and automating their estimation in an efficient, \emph{target-aware} manner.
This concept is realized by extending the PPS Turing to EPT using a combination
of program transformations and target-aware estimators.
We have shown that EPT estimates expectations effectively in practice, while its
modularity means that it can easily be built on by others.
Moreover, we believe the introduction of the high-level expectation programming concept can pave the way for exciting future advances.
While EPT focuses on the automation of TABI estimators, other implementations focusing on different approaches are conceivable---for example, systems targeting the automatic synthesis of control variates for a given input program---just as there are different PPSs focusing on distinct inference algorithms.

\begin{acknowledgements}
We would like to thank Sheheryar Zaidi for helpful discussions on 
configuring Annealed Importance Sampling. 
Tim Reichelt and Adam Golinski are supported by UK EPSRC CDT in Autonomous 
Intelligent Machines and Systems with the grants EP/S024050/1 (Tim Reichelt) and EP/L015897/1 (Adam Golinski).
Luke Ong is funded by EPSRC.
\end{acknowledgements}

\bibliography{references}

\onecolumn

\appendix

\section{Annealed Importance Sampling}
\label{apd:anis}

Annealed importance sampling (AnIS) \citepsupp{neal1998Annealed} is an inference algorithm which was
developed with the goal of efficiently estimating the normalization constant $Z$ of an 
unnormalized density $\gamma(x)$. It works by defining a sequence of annealing 
distributions $\pi_0(x), \dots, \pi_n(x)$ which interpolate between a simple 
base distribution $\pi_0(x)$ (typically the prior 
for a Bayesian model) and the 
complex target density $\pi_n(x) = \gamma(x)$. The most common scheme 
is to take
\begin{equation}
\pi_i(x) \propto \lambda_i(x) = \pi_0(x)^{1-\beta_n} \gamma(x)^{\beta_n},
\end{equation}
with $0\!=\!\beta_0\!<\!\dots\!<\!\beta_n\!=\!1$.
The algorithm further requires the definition of Markov chain transition 
kernels $\tau_1(x, x'), \dots, \tau_{n-1}(x, x')$ and proceed to generate 
the $j^{\text{th}}$ weighted sample as follows
First, sample initial particle $x^{(1)}_j \sim \pi_0(x)$, then
for $i = 1, \ldots, (n-1)$, generate $x^{(i+1)}_j \sim \tau_i(x^{(i)}_j, \cdot)$ and, finally, return sample $x^{(n)}_j$ with weight
\begin{equation}
    w_j = \frac{\lambda_1(x_j^{(1)})\lambda_2(x_j^{(2)}) \dots \lambda_n(x_j^{(n)})}{\pi_0(x_j^{(1)})\lambda_1(x_j^{(2)}) \dots \lambda_{n-1}(x_j^{(n)})}
\end{equation}
      
We can estimate expectations with the weights and samples just as in importance 
sampling.
Thus we can estimate the expectation and the normalization constant as 
\begin{align*}
\begin{split}
\mathbb{E}_{\pi(x)}[f(x)] \approx \frac{\sum_{j=1}^{N} w_j f(x^{(n)}_j)}{\sum_{j=1}^{N} w_j}\quad\text{and}\quad
Z \approx \frac{1}{N} \sum_{j=1}^{N} w_j.
\end{split}
\end{align*}

\subsection{Implementation Details of Turing Inference Engine}

The implementation of our new Turing inference engine is available at \url{https://github.com/treigerm/AnnealedIS.jl}.
It is a stand-alone package that can be used completely independently from EPT and is therefore useful for any Turing user who wishes to run AnIS on their model.
Furthermore, our implementation leverages the modularity of the Turing ecosystem by using existing MCMC transition kernels from the packages \texttt{AdvancedMH.jl} \citepsupp{turingdevelopmentteam2020TuringLang}
and \texttt{AdvancedHMC.jl} \citepsupp{xu2020AdvancedHMC}.

Keeping the same notation as above, given a Turing model the AnIS inference creates Julia functions for the prior density $\pi_0(x)$ and the unnormalized density $\gamma(x)$.
The unnormalized density of the program is evaluated as described in Section~\ref{sec:turing_details} and the prior density is evaluated similarly but ignores all the `likelihood' terms $h_j(y_j \mid \phi_j)$ and all the terms added with \ept{@addlogprob} primitive.
Once we have Julia functions for $\pi_0(x)$ and $\gamma(x)$ it is straightforward to create a function for the intermediate targets $\lambda_i(x)$ for a given $\beta_i$.
The Julia function for the intermediate targets $\lambda_i(x)$ can then be used by one of the MCMC samplers in \texttt{AdvancedMH.jl} or \texttt{AdvancedHMC.jl} to collect samples from the intermediate distributions.

\section{Theoretical Details}
\label{apd:theory_details}

\subsection{Assumptions in Definition~\ref{def:expectation_program}}
\label{sec:apd_assumptions}

To ensure correctness most PPSs assume that a particular inference 
algorithm will converge to the distribution of $F$ (i.e. the distribution over return values).
A standard PPS Monte Carlo inference engine will now produce a sequence of samples $F_n,~n=1,2,\dots$ and consistency requires that $F_n$ converges in distribution to $F$ as $n\to\infty$.
This is equivalent to requiring that for \emph{any} integrable function $h$, $\mathbb{E}[h(F_n)]\to \mathbb{E}[h(F)]$;
and it presupposes that the distribution of $F$ is a finite measure, i.e., $\mathbb{E}[F]$ is finite.
We thus see our assumption is strictly weaker than that of standard PPSs that allow return values from programs: we only need convergence in the case where $h$ is the identity mapping, not all integrable functions.

\subsection{Proof for Theorem~\ref{thm:tabi_valid}}

% \tabi*
\begin{restatable}{theorem}{tabi}
% \label{thm:tabi_valid}
Let $\mathcal{E}$ be a valid expectation program in EPT with unnormalized density $\gamma(x_{1:n})$, defined on possible traces $x_{1:n}\in\mathcal{X}$, with return value $F=f(x_{1:n})$.
Then
$\gamma_1^+(x_{1:n}):=\gamma(x_{1:n})\max(0,f(x_{1:n}))$, $\gamma_1^-(x_{1:n}):=-\gamma(x_{1:n})\min(0,f(x_{1:n}))$, and $\gamma_2(x_{1:n}):=\gamma(x_{1:n})$ are all valid unnormalized probabilistic program densities.
Further, if $\{\hat{Z}_1^+\}_{m}$, $\{\hat{Z}_1^-\}_{m}$, $\{\hat{Z}_2\}_{m}$ are sequences of estimators for $m \in \mathbb{N}^+$ such that
\vspace{-5pt}
\begin{align*}
\{\hat{Z}_1^\pm\}_{m} &\overset{p}{\to}
\int_\mathcal{X} \gamma^\pm_1(x_{1:n}) d\mu(x_{1:n}),
\\
\{\hat{Z}_2\}_{m} &\overset{p}{\to}
\int_\mathcal{X} \gamma_2(x_{1:n}) d\mu(x_{1:n})
\end{align*}
\vspace{-15pt}

where $\overset{p}{\to}$ means convergence in probability as $m\to \infty$, then
$(\{\hat{Z}_1^+\}_{m}-\{\hat{Z}_1^-\}_{m})/\{\hat{Z}_2\}_{m} \overset{p}{\to} \mathbb{E}[F].$
\end{restatable}
\begin{proof}
We start by noting that as $\gamma_2(x_{1:n})$ is identical to $\gamma(x_{1:n})$, it is by assumption a valid unnormalized program density. 
Meanwhile, by construction, $\gamma(x_{1:n})^+_1,\gamma(x_{1:n})^-_1 \geq 0, \forall x_{1:n} \in \mathcal{X}$.
Further, each can be written in the form of~\eqref{eq:ppl_density} by taking the correspond definition of $\gamma(x_{1:n})$ and adding in factors $\exp(\psi_{K+1})=\max(0,f(x_{1:n}))$ and $\exp(\psi_{K+1})=-\min(0,f(x_{1:n}))$ for $\gamma(x_{1:n})^+_1$ and $\gamma(x_{1:n})^-_1$ respectively.
To finish the proof that $\gamma^{\pm}(x_{1:n})$ are valid densities, we show that $0<Z_1^\pm<\infty$.

Starting with the standard definition of an expectation for arbitrary random variables, we can express $\mathbb{E}[F]$ as
\begin{align}
  \int_\mathcal{X} f(x_{1:n}) d\mathbb{P}(x_{1:n}) 
    &= \int_\mathcal{X} f^+(x_{1:n}) d\mathbb{P}(x_{1:n})  - \int_\mathcal{X} f^-(x_{1:n}) d\mathbb{P}(x_{1:n}). \label{eq:pos_neg_breakdown} \\
\intertext{
Noting that if $F$ is integrable then by the definition of the Lebesgue integral $\int_\mathcal{X} f^+(x_{1:n}) d\mathbb{P}(x_{1:n}) < \infty$ and $\int_\mathcal{X} f^-(x_{1:n}) d\mathbb{P}(x_{1:n}) < \infty$. Now inserting the distribution the program defines over $x_{1:n}$,}
    &= \int_\mathcal{X} f^+(x_{1:n}) \pi(x_{1:n}) d\mu(x_{1:n}) - \int_\mathcal{X} f^-(x_{1:n}) \pi(x_{1:n}) d\mu(x_{1:n}) \label{eq:dist_to_density} \\ 
\intertext{and noting that $\gamma(x_{1:n}) \geq 0$ for all $x_{1:n} \in \mathcal{X}$ and $0 < \int_\mathcal{X} \gamma(x_{1:n}) d\mu(x_{1:n}) < \infty$,}
    &= \frac{\int_\mathcal{X} f^+(x_{1:n}) \gamma(x_{1:n}) d\mu(x_{1:n}) - \int_\mathcal{X} f^-(x_{1:n}) \gamma(x_{1:n}) d\mu(x_{1:n})}{\int_\mathcal{X} \gamma(x_{1:n}) d\mu(x_{1:n})} \label{eq:unnormalized} \\
    &= \frac{\int_\mathcal{X} \gamma^+_1(x_{1:n}) d\mu(x_{1:n}) - \int_\mathcal{X} \gamma_1^-(x_{1:n}) d\mu(x_{1:n})}{\int_\mathcal{X} \gamma_2(x_{1:n}) d\mu(x_{1:n})} \label{eq:relabeling}
    =: \frac{Z^+_1 - Z^-_1}{Z_2}.
\end{align}
In our theorem statement we have assumed that $\{\hat{Z}_1^+\}_{m}\overset{p}{\to} Z_1^+$,
$\{\hat{Z}_1^-\}_{m}\overset{p}{\to} Z_1^-$, and
$\{\hat{Z}_2\}_{m}\overset{p}{\to} Z_2$, from which it now follows by Slutsky's Theorem that
\begin{align}
    \frac{\{\hat{Z}_1^+\}_{m}-\{\hat{Z}_1^-\}_{m}}{\{\hat{Z}_2\}_{m}} \overset{p}{\to} \frac{Z^+_1 - Z^-_1}{Z_2} = \mathbb{E}[F]
\end{align}
as required.
\end{proof}

\subsection{Details about Equation~(\ref{eq:ppl_density})}

Any probabilistic program defines a `density' function in the form of Equation~\eqref{eq:ppl_density}. 
This definition makes sense for a large class of programs, permitting branching on random variables, higher-order functions, recursion, stochastic memoization, and conditioning on internally sampled variables \citepsupp[\S4.3]{rainforth2017Automating}.
However, for this function to correspond to a valid unnormalized probability density we need to assume that a) the program halts with probability 1 and b) that the integral over the entire domain of $\gamma$ with respect to the implicitly defined reference measure is finite, i.e. $Z= \int_\mathcal{X} \gamma(x_{1:n}) d\mu(x_{1:n}) < \infty$ where $\mu$ is the reference measure and $\mathcal{X}$ denotes the space of valid program traces.

We further need to clarify our usage of the term `density function.'
In general, probabilistic programs denote measures (or kernels if there are free variables) \citepsupp{kozen1979semantics,staton2016semantics,borgstrom2011measure}.
When we talk about the density function of a probabilistic program, formally we are referring to the Radon-Nikodym derivative of the measure denoted by this program with respect to an appropriate reference measure, where this reference measure is itself implicitly defined by the program.

\section{Estimating Expectations in Turing}
\label{apd:turing_expectation}

\subsection{Standard approach}
\label{apd:turing_monte_carlo}

\begin{minted}[breaklines,escapeinside=||,mathescape=true,numbersep=3pt,gobble=2]{eptlexer.py:EPTLexer -x} 
@model function model(y=2)
    x |$\sim$| Normal(0, 1) 
    y |$\sim$| Normal(x, 1)
end

num_samples = 1000
posterior_samples = sample(model(), NUTS(0.65), num_samples)

f(x) = x^3
posterior_x = Array(posterior_samples[:x])
expectation_estimate = mean(map(f, posterior_x))
\end{minted}

Full example of the estimation of an expectation with the Turing language. 
The user first defines the model, then conditions it on some observed data, 
computes posterior samples and then uses these samples to compute a Monte Carlo
estimate of the expectation.

\subsection{Using generated quantities function}

When we designed the API Turing largely ignored the \ept{return} statements in 
the model definition. In the meantime Turing introduced a convenience function 
\ept{generated_quantities}. Given a model and $N$ samples it returns a list of 
the $N$ return values generated by running the program on each sample. 
Note that \ept{generated_quantities} reruns the entire \ept{model} function for 
each posterior sample to compute the return value. This means that for models 
which have an expensive likelihood computation the use of 
\ept{generated_quantities} might incur a significant overhead.

It is important to note that \ept{generated_quantities} is merely a convenience 
function and does not change how Turing interprets model definitions. In fact, 
the \ept{generated_quantities} function provides complimentary functionality and 
Turing models generated with EPT can use this function without problems.

The example from Section~\ref{apd:turing_monte_carlo} can be rewritten to use 
\ept{generated_quantities}:
\begin{minted}[breaklines,escapeinside=||,mathescape=true,numbersep=3pt,gobble=2]{eptlexer.py:EPTLexer -x} 
@model function model(y=2)
    x |$\sim$| Normal(0, 1) 
    y |$\sim$| Normal(x, 1)
    return x^3
end

num_samples = 1000
posterior_samples = sample(model(), NUTS(0.65), num_samples)

expectation_estimate = mean(generated_quantities(model(), posterior_samples))
\end{minted}

\section{Full Example of Macro Transformation}
\label{apd:macro_transformation}

The expectation
\vspace{-8pt}
\begin{minted}[breaklines,escapeinside=||,mathescape=true,numbersep=3pt,gobble=2,highlightlines={4}]{eptlexer.py:EPTLexer -x} 
@expectation function expt_prog(y)
    x |$\sim$| Normal(0, 1) 
    y |$\sim$| Normal(x, 1)
    return x^3
end
\end{minted}
\vspace{-8pt}
gets transformed into
\vspace{-8pt}
\begin{minted}[breaklines,escapeinside=||,mathescape=true,numbersep=3pt,gobble=2,highlightlines={4-8,14-18,24}]{eptlexer.py:EPTLexer -x} 
@model function gamma1_plus(y)
    x |$\sim$| Normal(0, 1) 
    y |$\sim$| Normal(x, 1)
    tmp = x^3
    if _context isa Turing.DefaultContext
        @addlogprob!(log(max(tmp, 0)))
    end
    return tmp
end

@model function gamma1_minus(y)
    x |$\sim$| Normal(0, 1) 
    y |$\sim$| Normal(x, 1)
    tmp = x^3
    if _context isa Turing.DefaultContext
        @addlogprob!(log(-min(tmp, 0)))
    end
    return tmp
end

@model function gamma2(y)
    x |$\sim$| Normal(0, 1) 
    y |$\sim$| Normal(x, 1)
    return x^3
end

expt_prog = Expectation(
    gamma1_plus,
    gamma1_minus,
    gamma2
)
\end{minted}
The type \ept{Expectation} is simply used to have one common 
object which stores the three different Turing models. Notice that for 
\ept{gamma2} the function body is identical to the original function.

For \ept{gamma1_plus} and \ept{gamma1_minus} we also have to check in what 
\ept{_context} the model is executed in. Turing allows to execute the model with 
different contexts which change the model behaviour. For example, there is a 
\ept{PriorContext} which essentially ignores the tilde statements which have 
observed data on the LHS. This is useful for evaluating the prior probability of 
some parameters. However, by default the \ept{@addlogprob} macro ignores the model 
context. As a consequence if a Turing model includes an \ept{@addlogprob} macro 
and is executed with a \ept{PriorContext} then it no longer calculates the log
prior probability but instead the log prior probability plus whatever value was 
added with the \ept{@addlogprob} statement. 
Since we want to use the Turing model with Annealed Importance Sampling 
we need to be able to extract the prior from our model and 
hence we need to ensure that we do not call \ept{@addlogprob} when executed in a \ept{PriorContext}. 
This is what the added \ept{if} clause ensures.

\section{Different Estimators for $Z_1^+$, $Z_1^-$ and $Z_2$}
\label{apd:positive_target_function}

The target function $f(x)=x^2$ in the following expectation is always positive:
\vspace{-8pt}
\begin{minted}[breaklines,escapeinside=||,mathescape=true,numbersep=3pt,gobble=2]{eptlexer.py:EPTLexer -x} 
@expectation function expt_prog(y)
    x |$\sim$| Normal(0, 1)
    y |$\sim$| Normal(x, 1)
    return x^2
end
\end{minted}
\vspace{-8pt}
Therefore, we already know that $Z_1^-=0$,
so it would be wasteful to spend computational resources on estimating $Z_1^-$. EPT
allows users to specify the marginal likelihood estimator for each of the terms in TABI
separately which means if the user knows that the target function is always 
positive they can specify that 0 samples should be used to estimate $Z_1^-$:
\vspace{-8pt}
\begin{minted}[breaklines,escapeinside=||,mathescape=true,numbersep=3pt,gobble=2]{eptlexer.py:EPTLexer -x} 
expct_estimate, diagnostics = estimate_expectation(
    expt_prog(2), TABI(
        TuringAlgorithm(AnIS(), num_samples=1000), # $Z_1^+$
        TuringAlgorithm(AnIS(), num_samples=0),    # $Z_1^-$
        TuringAlgorithm(AnIS(), num_samples=1000)  # $Z_2$
))
\end{minted}
\vspace{-8pt}
It is easy to see how this can be adapted to the case in which we have $Z_1^+=0$.
This interface is not just useful for avoiding unnecessary computation, in some 
cases the user might also want to have different marginal likelihood estimators 
for each term. 
This allows user to further tailor the inference algorithm for 
the given target function $f(x)$.

\section{Hyperparameters for Experiments}
\label{apd:exp_hyperparams}

EPT
runs standard annealed importance sampling twice: one time to estimate $Z^{+}_1$
and the other time to estimate $Z_2$. For each of the problems 
we always use the same hyperparameters for the annealed importance sampling 
algorithm both to run AnIS and for the two estimates in EPT.

\subsection{Posterior Predictive}

For the annealed importance sampling, we use a MH transition kernel with an 
isotropic Gaussian with covariance $0.5I$ as a proposal and $5$ MH steps on each 
annealing distribution. We use $100$ uniformly spaced annealing distributions.
For the MCMC, we collect $5\cdot10^7$ samples in total. To parallelise sampling 
we run $5\cdot10^3$ chains with $10^4$ samples each in parallel, 
discarding the first $10^3$ samples as burn-in. We use a MH transition kernel with 
standard normal proposal.

\subsection{SIR Model}
\label{apd:sir_hyperparamas}

For the annealed importance sampling estimators we use HMC 
transition kernels with a step size of 0.05, 10 leapfrog steps and 10 MCMC steps 
on each annealing distribution. We use 100 geometrically spaced annealing 
distributions.

For the MCMC model we collect $10^6$ samples in total with Turing's 
implementation of NUTS and a target acceptance rate of 65\%.\footnote{\url{https://turing.ml/dev/docs/library/\#Turing.Inference.NUTS}} 
We parallelise sampling over $10^2$ chains with $10^4$ samples and discard the
first $10^3$ samples as burn-in.

The ground truth is computed using importance sampling with $10^8$ samples and 
the prior as a proposal distribution. See Equation~\eqref{eq:bayesian_sir}
for the full SIR model including the priors.
The observed data was generated from the model described in \eqref{eq:bayesian_sir}
with $\beta = 0.25$, $I_0 = 100$, $N=10^4$ and $\phi = 10$ as the overdispersion parameter 
of the SIR model. We generate data for 15 time steps.

\subsection{Radon model}

We run EPT and AnIS with 200 intermediate distributions and one step of the 
dynamic HMC transition kernel \citepsupp{betancourt2018conceptual,hoffman2014TheNS} on 
each intermediate distribution with a step size of $0.044$. The step size was 
informed by running adaptive MCMC on the target distribution.

\section{SIR Experiment}
\label{apd:sir_exp}

We assume we are given data in the form of observations $y_i$, 
the number of observed newly infected people on day $i$. Fixing $\gamma=0.25$, this gives us the statistical model
\begin{subequations}
\label{eq:bayesian_sir}
\begin{align}
    \beta &\sim \text{TruncatedNormal}(2, 1.5^2, [0, \infty]),
    &I_0 &\sim \text{TruncatedNormal}(100, 100^2, [0, 10000]), \\
    S_0 &= 10000 - I_0, &R_0 &= 0,\\
    \mathbf{x} &= \texttt{ODESolve}(\beta, \gamma, S_0, I_0, R_0), 
    &y_i &\sim \text{NegativeBinomial}(\mu=x_i, \phi=0.5).
\end{align}
\end{subequations}
Here $\texttt{ODESolve}$ indicates a call to a numerical ODE solver 
which solves the set of equations~\eqref{eq:sir_diffeq}. It outputs $x_i$,
the predicted number of newly infected people on day $i$. We assume the observation process is noisy and model it using a negative binomial distribution, which is parametrised by a mean $\mu$ and 
an overdispersion coefficient $\phi$. 
For an in-depth discussion about doing Bayesian parameter inference in the SIR model we 
refer the reader to the case study of \citet{grinsztajn2020Bayesian}.

We are further given a cost function in terms of $R_0$, $\text{cost}(R_0) = 10^{12} * \text{logistic}(10R_0 - 30)$.
Intuitively, the cost initially increases exponentially with $R_0$. However, 
the total cost also saturates for very large $R_0$ (as the entire population becomes infected).

\section{Hierarchical Radon Model}
\label{apd:radon_exp}

The data for this problem was taken from: \url{https://github.com/pymc-devs/pymc-examples/blob/main/examples/data/radon.csv} (the repository uses an MIT license; the data contains no personally identifiable information).
The original data contains information about houses in 85 counties.
In order to make estimating normalization constants more tractable we reduce the 
number of counties to 20. 

Our target function is a function of predicted radon levels $y_i$ for a typical 
house with a basement (i.e. $x_i = 0$) in county $i$; $y_i$ is calculated using 
the predictive equation given in~\eqref{eq:radon_pred}.
We apply the function 
\begin{align*}
    f(y_i) = \frac{1}{1 + \exp(5 (y_i - 4))}
\end{align*}
to all the predicted radon levels and then take the product of all the $f_i$.
Finally, to avoid floating point underflow we set a minimum value of $1\mathrm{e}{-200}$.

\section{Multiple Expectations and Restrictions on $f(\cdot)$}
\label{apd:multiple_expectations}

The user is not restricted to defining only one expectation per model. 
By specifying multiple return values the user can specify multiple expectations.
The \ept{@exptectation} macro can recognise multiple return values and generates 
an expectation for each of them.
The user can then estimate each expectation independently using \ept{estimate_expectation}:
\vspace{-8pt}
\begin{minted}[breaklines,escapeinside=||,mathescape=true,numbersep=3pt,gobble=2]{eptlexer.py:EPTLexer -x} 
@expectation function expt_prog(y)
    x |$\sim$| Normal(0, 1)
    y |$\sim$| Normal(x, 1)
    return x, x^2, x^3
end
y_observed = 3
expt_prog1, expr_prog2, expt_prog3 = expt_prog
expct1 = expt_prog1(y_observed)
expct1_estimate, diagnostics = estimate_expectation(
  expct1, method=TABI(marginal_likelihood_estimator=TuringAlgorithm(
    AnIS(), num_samples=1000)))
\end{minted}

\section{Posterior Predictive Model in EPT}
\label{apd:post_pred_ept}

The expectation from Section~\ref{sec:exp_post_pre} can be defined 
in just 5 lines of code with EPT:
\vspace{-8pt}
\begin{minted}[breaklines,escapeinside=||,mathescape=true,numbersep=3pt,gobble=2]{eptlexer.py:EPTLexer -x} 
@expectation function expt_prog(y)
    x |$\sim$| MvNormal(zeros(length(y)), I)       # $\mathbf{x} \sim \mathcal{N}(\mathbf{x}; 0, I)$
    y |$\sim$| MvNormal(x, I)                      # $\mathbf{y} \sim \mathcal{N}(\mathbf{y}; \mathbf{x}, I)$
    return pdf(MvNormal(x, 0.5*I), -y)     |$\phantom{\sim}$|# $f(\mathbf{x}) = \mathcal{N}(-\mathbf{y}; \mathbf{x}, \frac{1}{2}I)$
end
\end{minted}

\section{Syntax Design}
\label{apd:syntax}

Prior works have considered two families of syntax design corresponding to the semantics required by EPT.
\citet{gordon2014Probabilistic} define the semantics for expectation computation 
via the syntax of probabilistic program's return expression, which is the approach we adopted in the design of EPT.
\citet{zinkov2017Composinga} take a different route and define the expectation semantics
via the use of syntax \mintinline{eptlexer.py:EPTLexer -x}{expect(m, f)} 
where \mintinline{eptlexer.py:EPTLexer -x}{m} is the program defining a measure and 
\mintinline{eptlexer.py:EPTLexer -x}{f} is the target function.

While designing the interface of EPT we considered two different design for 
defining the target function: either letting users specify the target function 
implicitly through the return values of the function or allowing users to 
specify a target function \ept{f} externally. The external function could then 
be passed to the \ept{estimate_expectation} function explicitly.

For EPT, we decided to adopt the former of the two designs mainly due to the simplicity of the resulting user interface and implementation.
In particular, it allows for simple to execute program transformations of 
the \mintinline{eptlexer.py:EPTLexer -x}{@expectation} macro 
into valid Turing programs to represent the individual densities, 
and thus the ability to use native Turing inference algorithms. 
Adopting the other approach
would additionally require designing and specifying the interface between 
the function signature \mintinline{eptlexer.py:EPTLexer -x}{f(.)} and 
the values of the named random draws performed by the model 
\mintinline{eptlexer.py:EPTLexer -x}{m}.
This would result in a more complex user-facing interface,
at the slight advantage of improved compositionality of models and functions.

\section{SIR Discussion}
\label{apd:sir_discussion}

\begin{figure*}[h!]
    \centering
    \begin{subfigure}{0.49\textwidth}
        \centering
        \includegraphics[clip,trim=1.0cm 0cm 0cm 0cm,width=\textwidth]{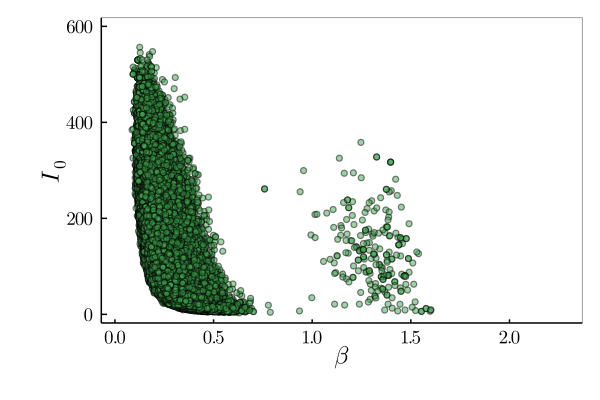}
        \vspace{-20pt}
        \caption{MCMC samples.}
        \label{fig:sir_mcmc}
    \end{subfigure}
    %\qquad
    \hfill
    \begin{subfigure}{0.49\textwidth}
        \centering
        \includegraphics[clip,trim=1.0cm 0cm 0cm 0cm,width=\textwidth]{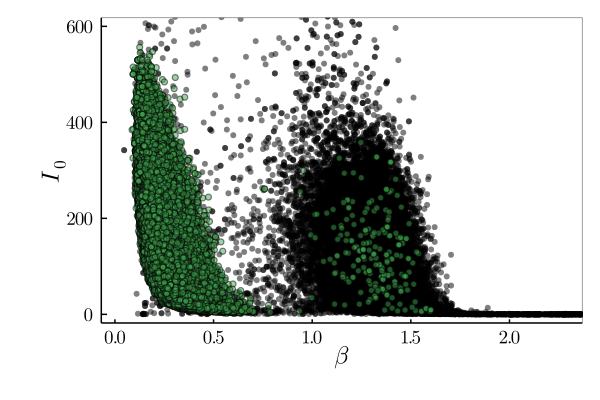}
        \vspace{-20pt}
        \caption{MCMC samples including burn-in samples (in black).}
        \label{fig:sir_mcmc_burn_in}
    \end{subfigure}
    \hfill
    \begin{subfigure}{0.49\textwidth}
        \centering
        \includegraphics[clip,trim=1.0cm 0cm 0cm 0cm,width=\textwidth]{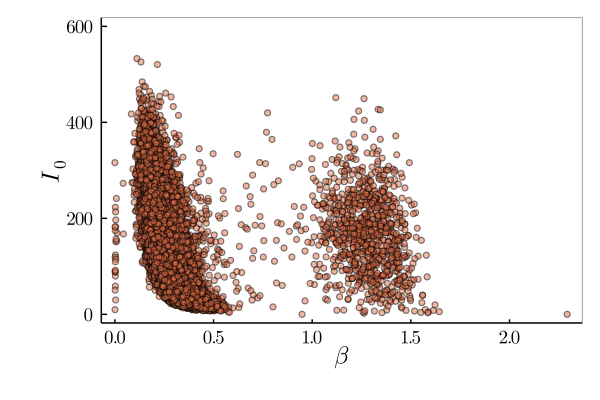}
        \vspace{-20pt}
        \caption{AnIS samples.}
        \label{fig:sir-Z2}
    \end{subfigure}
    \hfill
    \begin{subfigure}{0.49\textwidth}
        \centering
        \includegraphics[clip,trim=1.0cm 0cm 0cm 0cm,width=\textwidth]{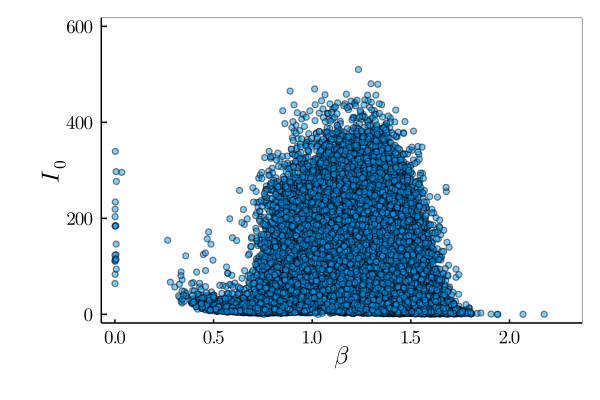}
        \vspace{-20pt}
        \caption{EPT samples for $Z_1$.}
        \label{fig:sir-Z1}
    \end{subfigure}
    % \vspace{-10pt}
    \caption{Samples from the different algorithms for the SIR model. Note that 
    for Figure~\ref{fig:sir_mcmc_burn_in} some burn-in samples lie outside the boundaries 
    of the plot but we adjusted the axis limits so that they are the same for all 
    plots to allow for easier comparison.}
	\vspace{-5pt}
    \label{fig:sir_samples}
\end{figure*}

In the SIR experiment AnIS achieved a significantly lower RSE than MCMC 
even though both are non-target-aware. 
Figure~\ref{fig:sir_samples} shows samples from the different algorithms.
The EPT samples for $Z_1$ visualise well in which regions of parameter space 
both the posterior and the target function have sufficient mass ($\beta \in [0.5, 2.0]$). 
The samples 
from AnIS and MCMC suggest that most of the posterior mass is located in the interval 
$\beta \in [0.3, 0.7]$. However, AnIS also generates a significant amount of 
samples in the parameter region $\beta \in [1.0, 1.5]$. The samples in this 
second ``mode'' are directly in the region of the target-aware samples. 
Further, the plots suggest that AnIS generates more samples in this regions than 
MCMC which is what allows AnIS to achieve a lower RSE. 
However, it seems that the AnIS represents the second ``mode'' disproportionally.
Specifically looking at the burn-in samples from 
MCMC in Figure~\ref{fig:sir_mcmc_burn_in} shows that MCMC will converge to the 
parameter space in $\beta \in [0.3, 0.7]$ even if the initial parameter samples 
are around $\beta \in [1.0, 1.5]$. This indicates that this is not a failure of MCMC 
to detect another mode but rather that there is 
negligible posterior mass in that parameter region.
Therefore the better performance of AnIS compared to MCMC 
seems to occur mostly because AnIS got lucky by accidentally generating samples in the 
right parameter region. 

\subsection{A Note on MCMC ESS}

The SIR experiment provides a good example of how the MCMC ESS \citepsupp{vehtari2020rank}
is unreliable for our use case. As detailed in Section~\ref{apd:sir_hyperparamas}
for MCMC we run $100$ chains with $10,000$ samples each. This is replicated $5$ 
times to get estimates on the variability in behaviour. 
After discarding the burn-in samples for each chain the $5$ replications give 
us the following final ESS estimates: $[631,360; \, 805,868; \, 873,269; \, 665,683; \, 5,114]$.
We observe that all but one replication give disproportionally high ESS estimates.
We found that the replication which gives a more conservative ESS estimate of $5,114$ is the 
replication which generated samples in the parameter region $\beta \in [1.0, 1.5]$
(see Figure~\ref{fig:sir_mcmc}). 
More importantly, the MCMC ESS estimates do not seem to show 
any correlation with the RSE values (see Figure~\ref{fig:sir_experiment}) which 
is the more important metric because it directly measures the error in our 
estimate. Therefore, we decided against using the MCMC ESS in our evaluation 
because it can give the impression that MCMC is performing well when it is actually 
failing dramatically (in terms of RSE).

\subsection{Additional Stan MCMC Baseline}

\begin{wraptable}{r}{0.6\linewidth}
    \vspace{-2pt}
	\caption{Quantiles of the RSE for different methods (the same performance metric as Figure~\ref{fig:sir_experiment}, left); computed over 5 runs.}
    \vspace{-10pt}
	\label{tab:additional_mcmc_baseline}
	\begin{center}
		\begin{small}
			\begin{sc}
				\begin{tabular}{llll}
					\toprule
					Method & 25\% Quantile & Median & 75\% Quantile \\
					\midrule
					EPT           & $2.96\mathrm{e}{-6}$ & $8.10\mathrm{e}{-6}$ & $2.92\mathrm{e}{-4}$ \\
					AnIS          & $0.02$ & $0.13$ & $0.15$ \\
					MCMC (Turing) & $0.96$ & $0.97$ & $0.97$ \\
					MCMC (Stan)   & $1.00$ & $1.00$ & $1.00$ \\
					\bottomrule
				\end{tabular}
			\end{sc}
		\end{small}
	\end{center}
	\vspace{-10pt}
\end{wraptable}

To validate our MCMC baseline we reimplemented the SIR model in Stan and used Stan's built-in default MCMC sampler.
We expressed the expectation within the \texttt{generated\_quantities} block leveraging the functionality described in Section~\ref{sec:related_work}.
We have picked Stan because its built-in MCMC sampler can be reasonably considered the state-of-the-art in its domain and has been extensively tested for correctness.
As shown in Table~\ref{tab:additional_mcmc_baseline}, Stan gives results that are similar to our current MCMC baseline (and potentially even a little worse). 
This demonstrates that the differences between existing PPSs are negligible compared to the effect of making inference target-aware.

\section{Effective Sample Size}

\begin{figure*}[h!]
    \centering
    \begin{subfigure}{0.32\textwidth}
        \centering
        \includegraphics[clip,trim=1.0cm 0cm 0cm 0cm,width=\textwidth]{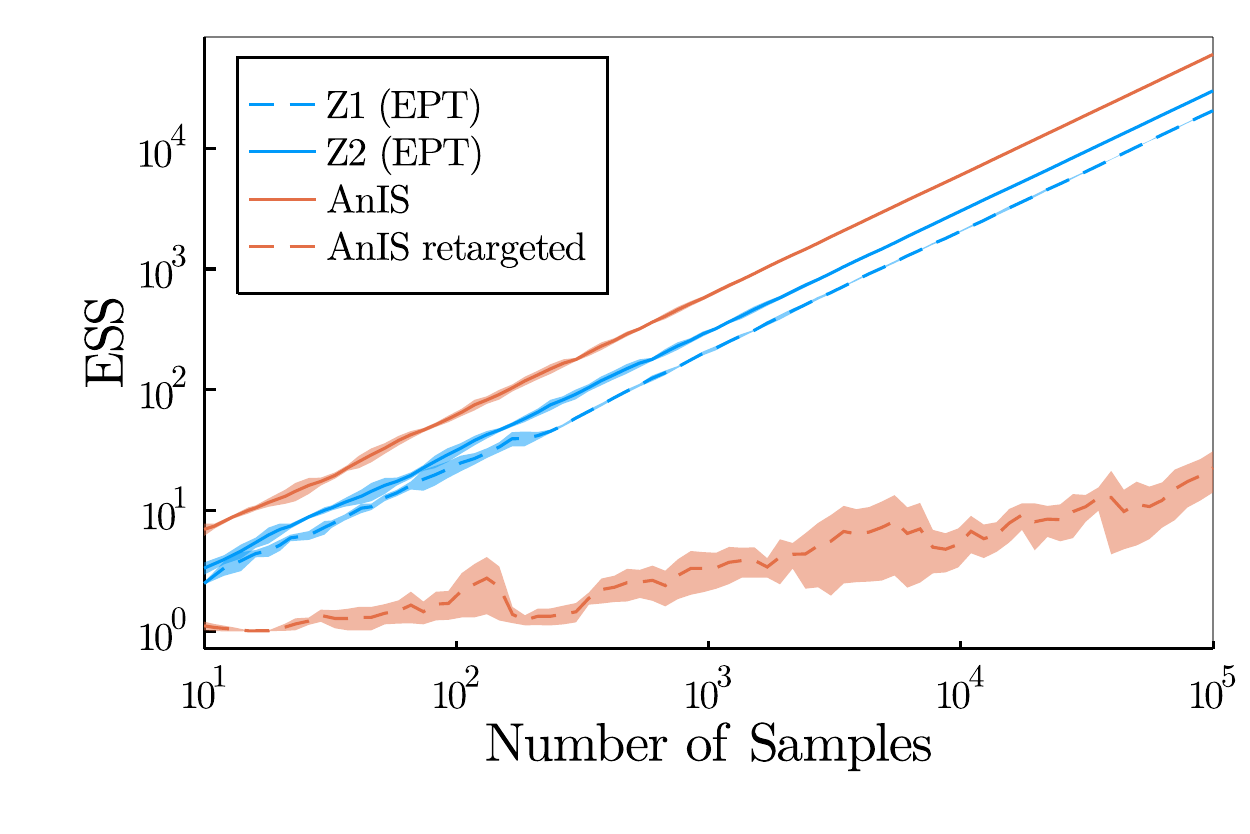}
        \vspace{-10pt}
        \caption{Gaussian Posterior Predictive.}
    \end{subfigure}
    %\qquad
    \hfill
    \begin{subfigure}{0.32\textwidth}
        \centering
        \includegraphics[clip,trim=1.0cm 0cm 0cm 0cm,width=\textwidth]{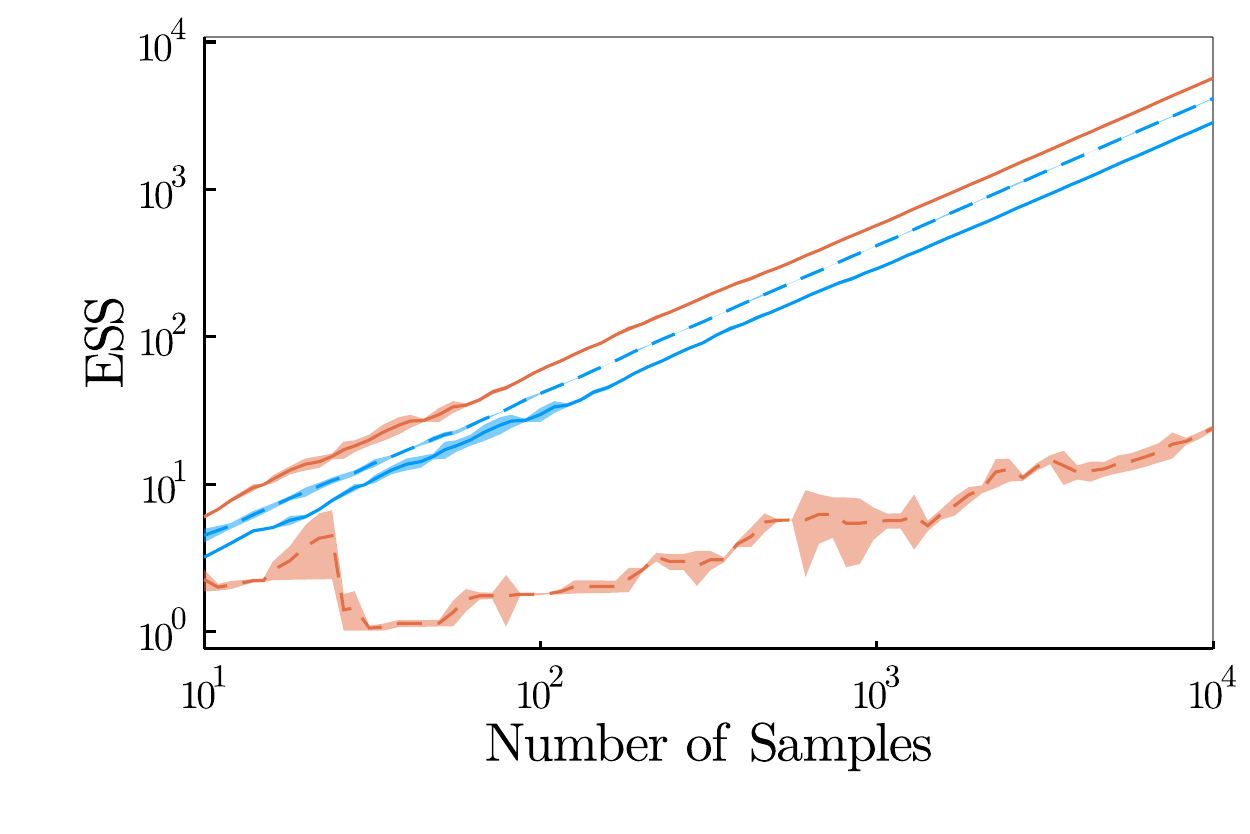}
        \vspace{-10pt}
        \caption{SIR.}
    \end{subfigure}
    \hfill
    \begin{subfigure}{0.32\textwidth}
        \centering
        \includegraphics[clip,trim=1.0cm 0cm 0cm 0cm,width=\textwidth]{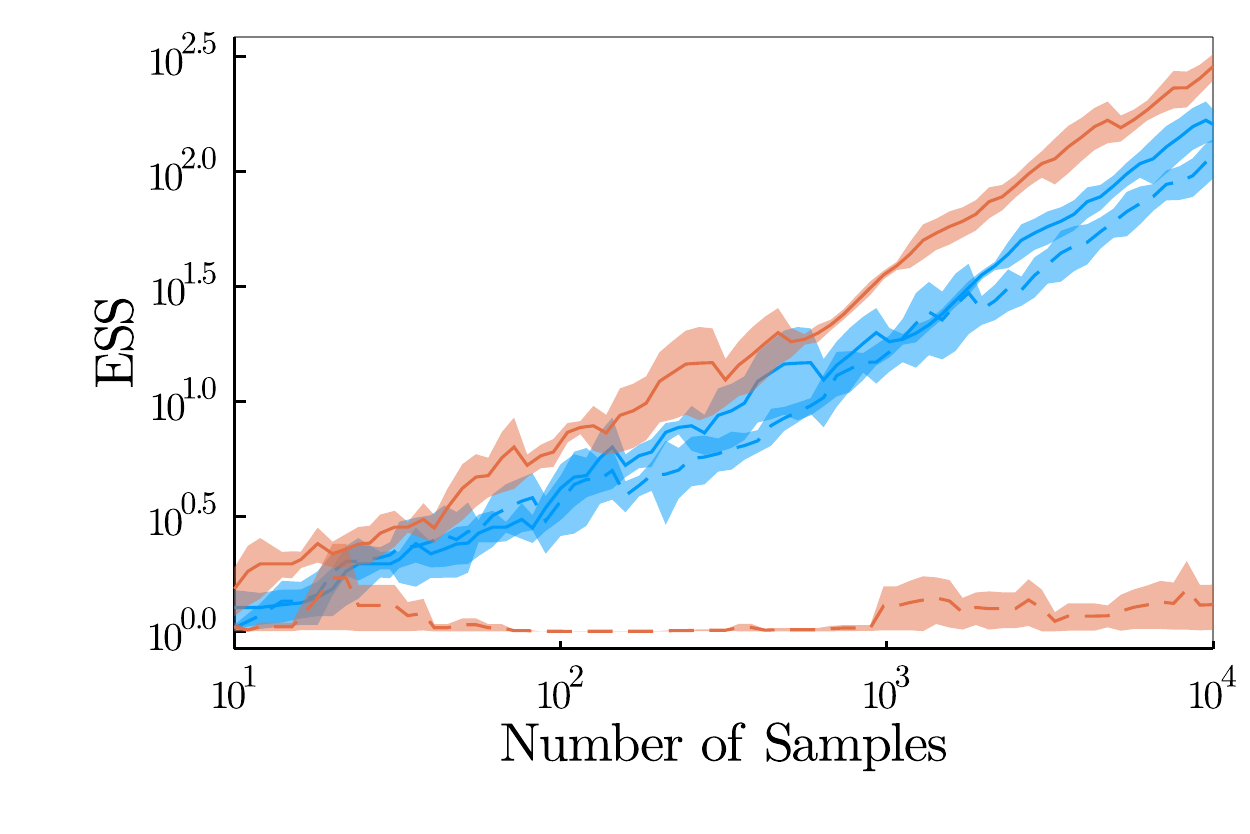}
        \vspace{-10pt}
        \caption{Radon.}
    \end{subfigure}
    % \vspace{-10pt}
    \caption{Individual ESS values as defined in Section~\ref{sec:experiments} for the
    three different experiments. Instead of taking $\min(\text{ESS}_{Z_1}, \text{ESS}_{Z_2})$ for EPT and 
    $\min(\text{ESS}_{Z_1}^{\text{AnIS}}, \text{ESS}_{Z_2}^{\text{AnIS}})$ for AnIS we 
    plot each value individually.}
	\vspace{-15pt}
    \label{fig:ess_details}
\end{figure*}

In Figure~\ref{fig:ess_details} we plot all the individual ESS values for EPT and the AnIS baseline.
Plotting each ESS value separately shows that the performance of AnIS is severely limited by its ability to generate samples in regions in which the target function $f(x)$ is large. 
This is indicated by the low values for $\text{ESS}_{Z_1}^{\text{AnIS}}$.

\section{Positive and Negative Target Functions}

\begin{figure*}[h!]
    \centering
    \begin{subfigure}{0.49\textwidth}
        \centering
        \includegraphics[clip,trim=1.0cm 0cm 0cm 0cm,width=\textwidth]{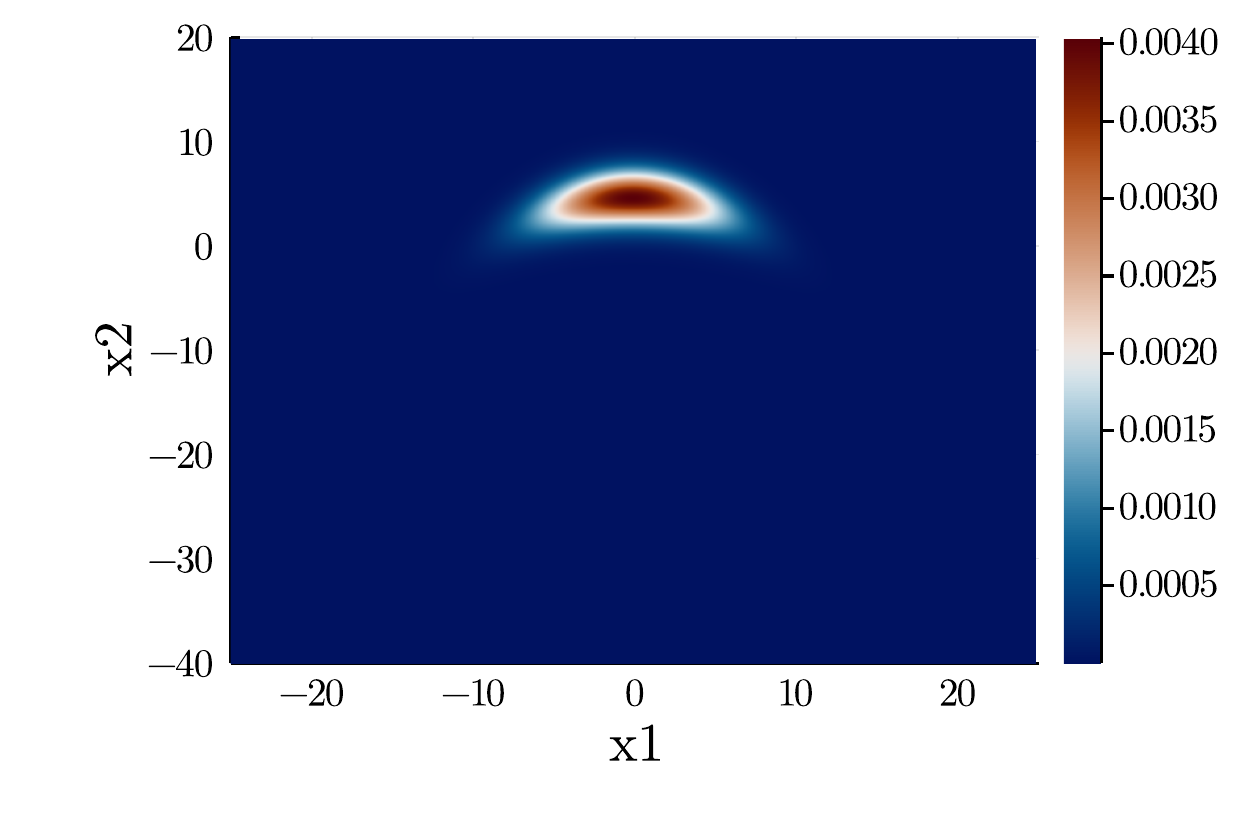}
        \label{fig:banana_density}
    \end{subfigure}
    %\qquad
    \hfill
    \begin{subfigure}{0.49\textwidth}
        \centering
        \includegraphics[clip,trim=1.0cm 0cm 0cm 0cm,width=\textwidth]{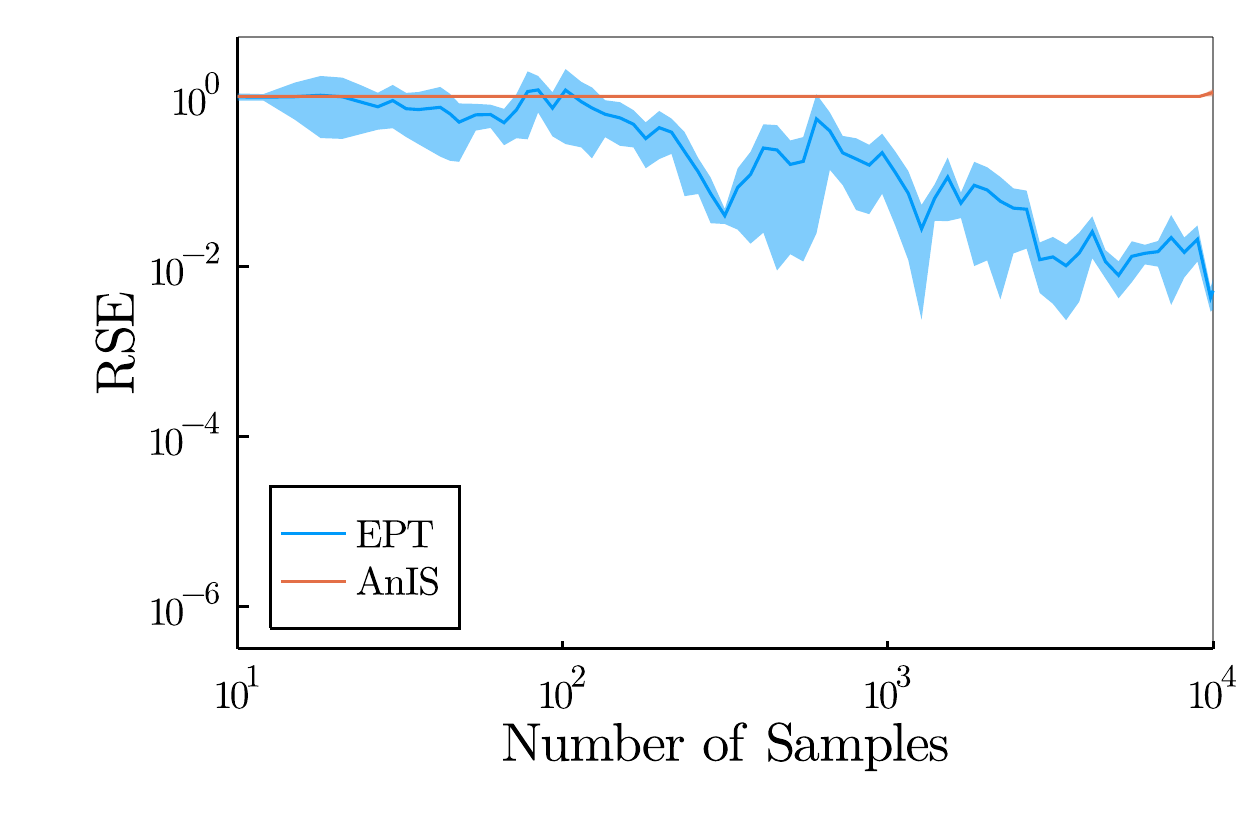}
        \label{fig:banana_error}
    \end{subfigure}
    \vspace{-20pt}
    \caption{Banana experiment. [Left] Heatmap of the density of the model. [Right] 
    Relative Squared Error for EPT and AnIS.}
	\vspace{-8pt}
    \label{fig:banana_experiment}
\end{figure*}

To demonstrate that EPT is also beneficial for target functions which are positive 
and negative we provide a brief description of a synthetic experiment.
We assume the following model which gives us a banana shaped density (see Figure~\ref{fig:banana_experiment}):
\vspace{-8pt}
\begin{minted}[breaklines,escapeinside=||,mathescape=true,numbersep=3pt,gobble=2]{eptlexer.py:EPTLexer -x}
@expectation function banana()
    x1 |$\sim$| Normal(0, 4)
    x2 |$\sim$| Normal(0, 4)
    @addlogprob!(banana_density(x1, x2))
    return banana_f(x1, x2)
end

banana_density(x1, x2) = -0.5*(0.03*x1^2+(x2/2+0.03*(x1^2-100))^2)
\end{minted}
\vspace{-8pt}
Note that there is no observed data in this experiment which is why we chose to 
express the banana distribution as an unnormalized density
(i.e. use the \ept{@addlogprob!} primitive). 
Our target function is given by
\vspace{-8pt}
\begin{minted}[breaklines,escapeinside=||,mathescape=true,numbersep=3pt,gobble=2]{eptlexer.py:EPTLexer -x}
function banana_f(x1, x2)
    cond = 1 / (1 + exp(50 * (x2 + 5)))
    return cond * (x1 - 2)^3
end
\end{minted}
\vspace{-8pt}
Note that the target function can be positive and negative. Figure~\ref{fig:banana_experiment}
shows the RSE for EPT and AnIS. We used an MH transition kernel and 200 
intermediate potentials for the Annealed Importance Sampling estimators. The RSE 
of AnIS does not improve because it fails to generate samples in the regions in 
which the target $f(x)$ is large. 
\cite{rainforth2020Target} provide a comparison to MCMC on a similar problem 
so we omit it here.

\bibliographysupp{references}
\bibliographystylesupp{plainnat}

\end{document}